\documentclass[10pt,twocolumn,letterpaper]{article}
\pdfoutput=1

\usepackage{cvpr}
\usepackage{times}
\usepackage{epsfig}
\usepackage{graphicx}
\usepackage{amsmath}
\usepackage{amssymb}
\usepackage{multirow}
\usepackage{booktabs} % tables
\usepackage{mathtools}
\usepackage[normalem]{ulem}
\usepackage{tabu}
\usepackage{subfigure}
\usepackage[ruled,linesnumbered]{algorithm2e}
\usepackage{xfrac}
\usepackage{caption}
\usepackage{color}
\usepackage{dblfloatfix}

\newcommand\Set[2]{\{\,#1\mid#2\,\}}
\definecolor{blueviolet}{RGB}{55,0,140}
\definecolor{purple}{RGB}{100,0,156}
\definecolor{darkergreen}{RGB}{10,180,20}

\newcommand{\beginsupplement}{%
        \setcounter{table}{0}
        \renewcommand{\thetable}{S\arabic{table}}%
        \setcounter{figure}{0}
        \renewcommand{\thefigure}{S\arabic{figure}}%
     }

\makeatletter
\newcommand\refwithdefault[2]{%
  \@ifundefined{r@#1}{%
    #2%
  }{%
    \ref{#1}%
  }%
}
\makeatother

\newcommand{\comment}[1]{}

\definecolor{orange}{RGB}{255,127,0}

\usepackage[pagebackref=true,breaklinks=true,letterpaper=true,colorlinks,bookmarks=false]{hyperref}

\cvprfinalcopy % *** Uncomment this line for the final submission

\def\cvprPaperID{305} % *** Enter the CVPR Paper ID here

\ifcvprfinal\fi

\graphicspath {{./figures/}}
\begin{document}

%%%%%%%%% TITLE

\title{
Keep it Unreal:\\
Bridging the Realism Gap for 2.5D Recognition with Geometry Priors Only
}

\makeatletter
    \renewcommand*{\@fnsymbol}[1]{\ensuremath{\ifcase#1\or $CO$\or \dagger\or \ddagger\or
        \mathsection\or \mathparagraph\or \|\or **\or \dagger\dagger
        \or \ddagger\ddagger \else\@ctrerr\fi}}
    \makeatother
\author{
%ArXiv Submission\\
%Siemens Corporate Technology, Germany --
%Siemens Corporate Technology, USA\\
%Technical University of Munich, Germany --
%University of Passau, Germany
Sergey Zakharov\thanks{These authors contributed equally to the work.}\hspace{1em}$^{,1}$, Benjamin Planche\footnotemark[1]\hspace{1em}$^{,1}$, 
Ziyan Wu$^{2}$, Andreas Hutter$^{1}$, Harald Kosch$^{3}$, Slobodan Ilic$^{1}$\\
$^{1}$Siemens Corporate Technology, Germany\\
{\tt\small \{sergey.zakharov, benjamin.planche, andreas.hutter, slobodan.ilic\}@siemens.com}
\and
$^{2}$Siemens Corporate Technology, USA\\
{\tt\small ziyan.wu@siemens.com}
\and
$^{3}$University of Passau, Germany\\
{\tt\small harald.kosch@uni-passau.de}
}

\maketitle
%\thispagestyle{empty}

%%%%%%%%% ABSTRACT
\begin{abstract}
With the increasing availability of large databases of 3D CAD models, depth-based recognition methods can be trained on an uncountable number of synthetically rendered images.
However, discrepancies with the real data acquired from various depth sensors still noticeably impede progress.
Previous works adopted unsupervised approaches to generate more realistic depth data, but they all require real scans for training, even if unlabeled. This still represents a strong requirement, especially when considering real-life/industrial settings where real training images are hard or impossible to acquire, but texture-less 3D models are available.
We thus propose a novel approach leveraging only CAD models to bridge the realism gap. Purely trained on synthetic data, playing against an extensive augmentation pipeline in an unsupervised manner, our generative adversarial network learns to effectively segment depth images and recover the clean synthetic-looking depth information even from partial occlusions.
As our solution is not only fully decoupled from the real domains but also from the task-specific analytics, the pre-processed scans can be handed to any kind and number of recognition methods also trained on synthetic data.
Through various experiments, we demonstrate how this simplifies their training and consistently enhances their performance, with results on par with the same methods trained on real data, and better than usual approaches doing the reverse mapping.

%\keywords{realism gap, domain adaptation, depth recognition}
\end{abstract}

\begin{figure}
  \centering
  \includegraphics[width=1\linewidth]{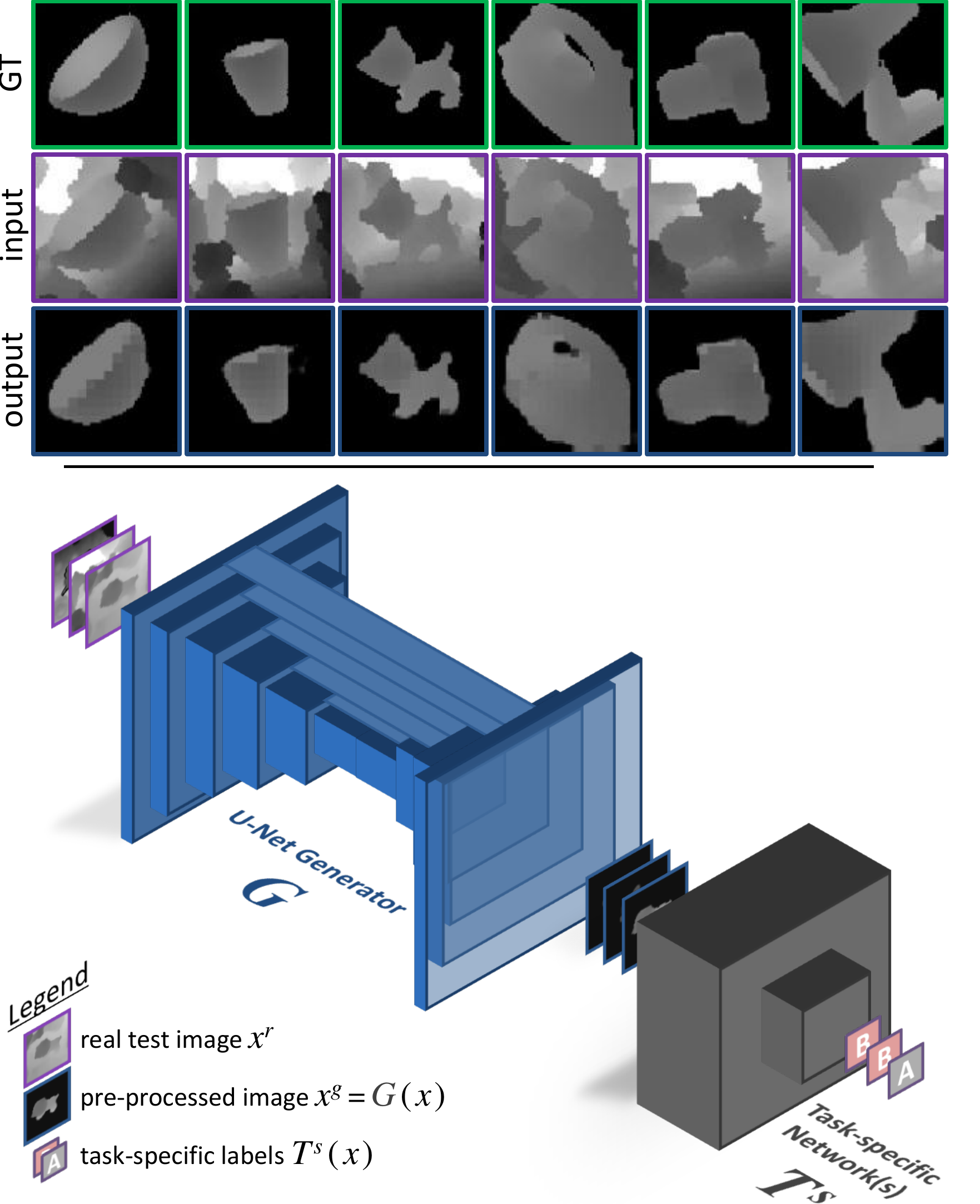}
  \caption{
  \textbf{Usage and results.} Trained on augmented data from 3D models, our network $G$ can map real scans (\textit{input}, here from LineMOD~\cite{hinterstoisser2012model}) to the synthetic domain (\textit{output}, compared to ground-truth \textit{GT}). The pre-processed data can then be handed to various recognition methods ($T^s$) to improve their performance.
  }

  \label{fig:pipeline-use}
\end{figure}

%%%%%%%%% BODY TEXT
\section{Introduction}
% !TeX spellcheck = en_US

\begin{figure*}[t]
  \centering
  \includegraphics[width=1\linewidth]{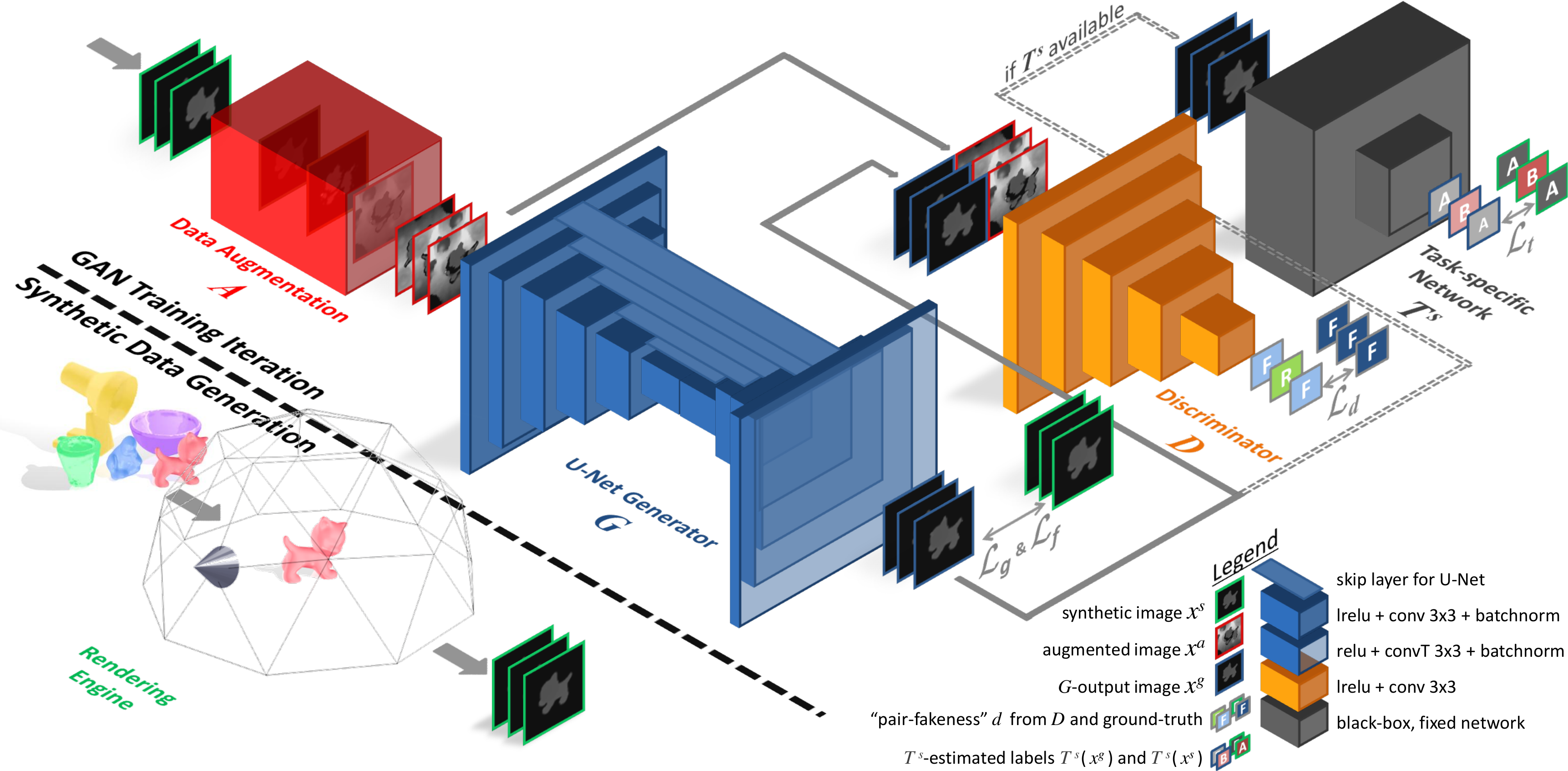}
  \caption{\textbf{Training of the depth-processing network $G$}. Following the conditional GAN architecture, a generator $G$ is trained against a discriminator $D$ to recover the original noiseless image from a randomly augmented, synthetic one. Its loss combines similarity losses $\mathcal{L}_{g}$ and $\mathcal{L}_{f}$, the conditional discriminator loss $\mathcal{L}_{d}$, and optionally a feature-similarity loss $\mathcal{L}_{t}$ if a task-specific method $T^s$ is provided at training time (network architectures are detailed in Section \ref{sec:archi}).}
  \label{fig:pipeline}  
\end{figure*}

Recent progress in computer vision has been dominated by deep neural networks trained over large datasets of annotated data. Collecting those is however a tedious if not impossible task. In practice, and especially in the industry, 3D CAD models are widely available, but access to real physical objects is limited and often impossible (\eg one cannot capture new image datasets for every new client, product, part, environment, \etc.). Moreover, appearance of industrial objects changes during their use, which naturally imposes the use of depth sensors as a logical choice for robust recognition.
For these reasons, numerous recent studies started using synthetic depth images rendered from databases of pre-existing 3D models~\cite{nips12:d3d,cviu15:synt,nips12:crdl,iccv13:fpe,cvpr15:sund,icra14:bb,carlucci2016deep}.
With no theoretical upper bound on generating data to train complex models for recognition~\cite{iccv15:mvcnn,iros15:vn,cvpr15:3dsp,icra09:l3d} or fill large databases for retrieval tasks~\cite{accv10:els,eccv14:rtf}, research continues to gain impetus in this direction. 

However, the performance of these approaches is often affected by the {\em realism gap}, \ie the discrepancy between the synthetic depth (2.5D) data the recognition methods are learning from and the more complex real data they have to face afterwards.
Some approaches address this matter by generating datasets in such a way that they mimic captured ones, \eg using sensor simulation methods~\cite{keller2009real,gschwandtner2011blensor,landau2015simulating,landau2016,planche2017depthsynth}.
This is however a non-trivial problem, as it is extremely difficult to account for all the physical variables affecting the sensing process. 
This led to another recent trend that tries to teach neural networks a direct mapping from synthetic to real images in order to generate pseudo-realistic training data~\cite{li2017triple,isola2016image,taigman2016unsupervised,shrivastava2016learning,bousmalis2016unsupervised}, or tries to force the task-specific methods to learn domain-independent features.
But such methods require real images for their training, circling back to the original problem. 
One could also wonder if ``tarnishing'' noiseless synthetic data, losing clarity and information for realism, is the best way to tackle the problem (see supplementary material for a clear visualization of those drawbacks).
Would not the reverse process be more beneficial, \ie projecting complex data into the simulated, synthetic domain we fully have control over?

Following this philosophy and considering the aforementioned industrial use-cases, we introduce a novel approach (\textit{UnrealDA}) to bridge the realism gap by mapping unseen real images toward the synthetic image domain used to train recognition methods, thus greatly improving their performance. Our contributions can be summarized as follows:

\par\noindent
\textbf{(a) Depth data segmentation and recovery} --  To the best of our knowledge, we propose the first end-to-end framework for depth image denoising and segmentation based on a custom GAN method purely trained on synthetic data \ie with only CAD models as prior. 
Not only our \textit{UnrealDA} method recognizes and segments the target objects out of real scans, but it can also partially recover missing information \eg caused by occlusions or sensing errors (as shown in Figure~\ref{fig:pipeline-use}). %, further facilitating any recognition process afterwards.

\par\noindent 
\textbf{(b) Independence from recognition methods} -- Most domain adaptation methods re-train recognition networks or constrain their architecture. Our solution converts real scans into denoised uncluttered versions which can be handed over to any task-specific networks already trained on noiseless synthetic data, with no structural adaptation needed. 

\par\noindent
\textbf{%(a) \textbf{Uncoupling from target real domains} --
(c) Extensive data augmentation} -- 
Domain adaptation approaches usually require to have access to images from the target domain. Completely decoupled from target real domains, \textit{UnrealDA} is trained on synthetic data only, generated from the 3D models of the target objects, and generalizes well to the real environments these objects can be found in. We leverage an extensive data augmentation pipeline to teach our solution a mapping from any kind of tampered images to their noise-free version.
\par\noindent
\textbf{(d) Performance improvement} --  Our solution considerably improves the performance on real data for algorithms pre-trained on noiseless synthetic data. More importantly it offers better results when compared to the same methods directly trained on images generated with our augmentation pipeline. Performance even compares to solutions trained on a subset of real images from the target domain and outperforms related works doing the opposite \ie generating realistic images from synthetic renderings.
By decoupling data augmentation for domain invariance and feature learning, our pipeline thus makes recognition methods easier to train, and overall more effective. 

\par\noindent
\textbf{(e) Interpretability} -- Similarly to~\cite{taigman2016unsupervised,shrivastava2016learning,bousmalis2016unsupervised}, our method adapts images, so its results can be easily interpreted (unlike adapted weights or features for instance).

The rest of the paper is organized as follows. In Section~\ref{sec:rw}, we 
provide a survey of pertinent work to the reader. We then introduce our framework and its components in Section~\ref{sec:mth}. Section~\ref{sec:exp} shows the effectiveness and flexibility of our tool, pairing it with state-of-the-art methods for various tasks, before concluding in Section~\ref{sec:cnc}.

%-------------------------------------------------------------------------
\section{Related Work}
\label{sec:rw}
% !TeX spellcheck = en_US

With the popular advocacy of structured-light sensors for vision applications, depth information became the support of active research. However, compared to the synthetic data generated for the training of many deep learning methods, real scans contain clutter and occlusions, and are corrupted by noise varying from one sensor to another.
In this section, we present recent recognition approaches employing synthetic scans for their training, and develop on the current trends to deal with the discrepancy challenge.

%\cSlo{became widely available to Computer Vision researchers. Consequently, 3D object recognition and pose estimation in depth images with the
%means of deep learning requires access to a large number of 3D depth data. With the availability of 3D models such depth data can be obtained by synthetic renderings, but they are quite different from the real depth data acquired from the sensors. Real depth images contain background clutter and occlusions, they are corrupted by sensor noise being different for every sensor type and therefore making realistic images 
%from the synthetic ones remains to be a challenge. } 
%is the support of active research within this domain. We emphasize on recent recognition approaches employing synthetic scans for their training, and develop on the current trends to deal with the discrepancy between such data and real target domains.

\par\noindent
\textbf{3D Object Recognition in Depth Images:}
Current state-of-the-art methods in visual recognition (\eg classification, pose estimation, segmentation, \etc) are view-based, \ie working on multiple 2D views covering the object's surface. A popular representative is the algorithm of Wohlhart~\etal~\cite{Wohlhart15}, where images are mapped to a lower dimensional descriptor space in which objects and their poses are separated. This mapping is learned by a convolutional neural network (CNN) using a triplet loss function, which pulls similar samples together and pushes dissimilar ones further away from each other. 
During the test phase, the trained CNN is used to get the descriptor of an unseen patch to find its closest neighbors in a database of pre-stored descriptors for which poses and classes are known. For some experiments in this paper, we employ a more recent work by Zakharov~\etal \cite{zakharov2017} which extends the method by including in-plane rotations and introduces an improved loss function.
Such deep learning methods however need large amounts of data for their training, which are extremely tedious to collect and accurately label (especially when 3D poses are considered for ground truth).

\par\noindent
\textbf{Bridging the Realism Gap:}	
Renewed efforts were put into the augmentation of existing datasets (\eg by applying noise or rendering unseen images~\cite{rematas2014image,ax15:rcnn}), or into the generation of purely synthetic datasets~\cite{cvpr15:3dsp,iros15:vn,iccv15:mvcnn,ax15:rcnn} (\eg leveraging recent 3D model databases~\cite{cvpr15:3dsp,shapenet2015}) to train more flexible estimators. 
As previously mentioned, the differences between the synthetic domain they are trained on and the real domain they are applied to still heavily affects their accuracy.
A first straightforward approach to reduce this discrepancy is to improve the quality of the sensor output, \eg by partially compensating the sensor noise~\cite{jungnoise,lu2014depth,liu2016computationally,liu2017efficient}, filling some of the missing data~\cite{liu2016computationally,liu2017efficient}, or improving the overall resolution~\cite{xie2016edge}. 
Common approaches employ filtering methods (\eg Gaussian, anisotropic, morphological, or learned through deep learning)~\cite{jungnoise,liu2016computationally,xie2016edge,zhang2016fast,zhu2017image,gu2017learning} and/or other sensing modalities (\eg aligned RGB data) to recover part of the depth information~\cite{lu2014depth,zhang2016fast,zhu2017image,gu2017learning}.
Though effective at denoising, these methods can only partially declutter the images and cannot make up for missing information. 

%\cSlo{I would remove this paragraph and keep the previous one.}
%Besides simple corrections, other image pre-processing procedures can be applied to help recognition \eg by handing them normalized data~\cite{krizhevsky2012imagenet,simonyan2014very}, enhanced content (\eg edge or contrast enhancements)~\cite{cirecsan2012multi,krig2014computer}, additional modalities (\eg normals computed from depth data)~\cite{nips12:crdl,zakharov2017}, segmented patches~\cite{dinh2014image,toshev2014deeppose,girshick2014rich}, \etc. Still, similarly to features hand-crafting, efforts put on data pre-processing for machine learning declined with the success of deep learning. \cBen{remove/shorten previous paragraph?}

%Currently, most of the research works tackling the \textit{realism gap} focus instead either on improving the realism of the training data, or enforcing the recognition methods to learn features invariant to the discrepancy.

Solutions tackling the \textit{realism gap} by improving the realism of the training data are divided into two complementary categories. While some researchers are coming up with more advanced simulation pipeline~\cite{keller2009real,gschwandtner2011blensor,landau2015simulating,landau2016,planche2017depthsynth} to generate realistic images by imitating the sensors mechanisms and taking into account environmental attributes; 
others are trying to learn a mapping from synthetic to real image domain using conditional GANs~\cite{li2017triple,isola2016image,taigman2016unsupervised,shrivastava2016learning,bousmalis2016unsupervised} or CNN style transfer methods~\cite{gatys2015neural,gatys2016image}.
The former solutions however need precise sensor models and object representations (\eg reflectance models) for the simulation to properly work; and even if unsupervised, the latter methods require real data to learn the target distribution. 
The same goes for unsupervised domain adaptation methods~\cite{csurka2017domain} trying to force recognition methods to learn domain-invariant features~\cite{tzeng2014deep,tzeng2017adversarial,ganin2015unsupervised,ganin2016domain}, or to learn a mapping  from the target image domain back to the source one~\cite{taigman2016unsupervised}: they all need real data (even if unlabeled) for their training.

Sadeghi and Levine~\cite{sadeghi2016cad} as well as Tobin \etal~\cite{tobin2017domain} managed to successfully train complex recognition methods by adding enough variability to the rendered data (different textures, lighting conditions, scene composition, \etc) so that the methods learn domain-invariant features. We apply the same principle, but to train a generative adversarial network able to adapt images from any pseudo-realistic domain to the noiseless synthetic one. Any recognition methods themselves trained on noiseless data can then be plugged to our pipeline with no need for any fine-tuning nor changes.

\par\noindent
\textbf{Generative Adversarial Networks (GANs):}
Introduced by Goodfellow et al.~\cite{goodfellow2014generative}, and quickly improved and derived through numerous works, \eg~\cite{radford2015unsupervised,salimans2016improved,li2016precomputed,isola2016image,yu2017seqgan}, 
the GAN framework has proven itself a great choice for image generation~\cite{goodfellow2014generative,radford2015unsupervised,chen2016infogan}, 
edition~\cite{isola2016image,zhu2017unpaired,shrivastava2016learning,bousmalis2016unsupervised}, 
or segmentation~\cite{luc2016semantic,nie2017medical,xue2017segan}. 
The generator network in these solutions benefits from competing against a discriminator one (with adversarial losses) to properly sample sharp, realistic images from the learned distribution. 
Methods conditioned on noise vectors~\cite{gauthier2014conditional,denton2015deep}, 
labels~\cite{gauthier2014conditional,liu2016coupled,chen2016infogan} 
and/or images~\cite{isola2016image,zhu2017unpaired,liu2016coupled,shrivastava2016learning,bousmalis2016domain,bousmalis2016unsupervised} soon appeared to add control over the generated data.
Given these additions, recognition pipelines started integrating conditional GANs. Some works are for instance using a classifier network along their discriminator, to help the generator grasp the conditioned image distribution by back-propagating the classification results on generated data~\cite{li2017triple,bousmalis2016unsupervised}; while others are using GANs to estimate the target domain distribution, to sample training images for their classifier~\cite{ganin2015unsupervised,ganin2016domain,taigman2016unsupervised,shrivastava2016learning,bousmalis2016domain,bousmalis2016unsupervised}.
% G learns to fool D (generating realistic images) while preserving the task-specific label estimated by T
% T learns to classify on images generated by G similar to those of the target domain (domain adaptation) or images generated by G from various domain (domain invariance)

The GAN our pipeline uses is based on the architecture by Isola~\etal~\cite{isola2016image}, augmented with foreground-similarity and task-specific losses similar to those introduced by Bousmalis~\etal~\cite{bousmalis2016domain,bousmalis2016unsupervised}. 
Their method however requires unlabeled target data and trains a classifier network both on source and augmented data; while our work assumes no access to target data at all, and only optionally uses a fixed recognition network to train the GAN. This makes our solution faster to train and much easier to deploy.

%-------------------------------------------------------------------------
\section{Methodology}
\label{sec:mth}
% !TeX spellcheck = en_US

Driven by the idea of learning from synthetic depth scans for recognition applications, we developed a method that brings real test images close to the noiseless uncluttered synthetic data the recognition algorithms are used to.
%We present our pipeline to pre-process real images to bring them closer to the noiseless uncluttered synthetic data used to train the recognition methods, making it easier for them to apply what they learned during training. Purely trained on synthetic data generated from the 3D CAD models of the target objects, our method is able to denoise the real images, segment the detected objects out of them, and partially recover missing parts.
Itself trained on a synthetic dataset rendered from the 3D CAD models of the target objects, our \textit{UnrealDA} solution is able to denoise the real images, segment the objects, and partially recover missing parts. Inspired by previous works on GAN-based image generation, and making use of an extended data augmentation process for its training (as shown in Figure~\ref{fig:pipeline}), our pipeline is both straightforward and effective.

Formalizing our problem, let $X^s_c=\Set{x^s_{c,i}}{\forall i \in N^s_c}$ be a dataset made of a large number $N^s_c$ of noiseless synthetic depth scans $x^s_c$, rendered from the 3D model of the class $c$; and let $X^s=\Set{X^s_c}{\forall c \in C}$ be the synthetic dataset covering all object classes $C$. We similarly define $X^r$ a dataset of real images unavailable at training time.

Finally, let \hbox{$T^s(x\ ; \theta_T)\to \widetilde{y}$} be any recognition algorithm which given a depth image $x$ returns an estimate $\widetilde{y}$ of a task-specific label or feature $y$ (\eg class, pose, mask image, hash vector, \etc). We suppose $T^s$ is pre-trained on synthetic scans covering $C$ and their labels. Its parameters $\theta_T$ are considered fixed \ie at no point do we alter its architecture, trained weights, \etc. We consider both cases when $T^s$ is set and provided for the training of $G$, and when it is not.

Given this setup, we present how our pipeline trains a function $G$ purely on synthetic data $X^s$ (and thus in an unsupervised manner), to pre-process tampered or cluttered images of $C$ instances, to consistently increase the probability that $T^s(G(x^r)) = \widetilde{y^g}$ is accurate compared to $T^s(x^r) = \widetilde{y^r}$, given $T^s$ also trained on synthetic data (\cf Figure~\ref{fig:pipeline-use}). Formally, if $y^r$ is a true, unknown label of $x^r$; then we want:
\begin{equation}
\Pr\left(\widetilde{y^g} = y^r \mid x^r \in X^r\right) > \Pr\left(\widetilde{y^r} = y^r \mid x^r \in X^r\right)
\label{eq:proba_accu}
\end{equation}
%\cSlo{indicating thet we want to achieve higher probability of the classification with generator produced images then with the real one. This is true since our recognition method $T^s$ is trained on synthetic images.}
To achieve this, we describe how $G$ is trained against a data augmentation pipeline \hbox{$A(x^s, z) \to x^a_z$}, with $z$ a noise vector randomly defined at every training iteration and $x^a_z$ the augmented image (as shown in Figure~\ref{fig:augmentation_and_results}). We demonstrate how given a complex and stochastic procedure $A$, the mapping learned by $G$ transposes to the real images.

\subsection{Unsupervised Learning from Synthetic Data Only} \label{sec:learning_process}

%\begin{figure}[t]
%  \centering
%  \includegraphics[width=1\linewidth]{figures/augmentation_and_training_results}
%  \caption{\textbf{Augmentation and Training Results}. Effects of the main augmentation methods (FG - foreground distortion; BG - background noise; OC - occlusions) for different amplitudes or types of noise; and qualitative training results (last two columns).}
%  \label{fig:pipeline-use}  
%\end{figure}

Following recent works in domain adaptation~\cite{li2017triple,isola2016image,taigman2016unsupervised,shrivastava2016learning,bousmalis2016unsupervised}, we adopt a generative adversarial architecture.
We define first a generator function \hbox{$G(x\ ; \theta_G)\to x^g$}, parametrized by a set of hyper-parameters $\theta_G$ and conditioned by an image $x$ to generate a version $x^g$.
During training, the task of $G$ is to restore the noiseless depth data from its augmented version, \ie to obtain $x^g \simeq x^s$ given $x = A(x^s, z)$.
Then, provided an image $x \in X^r$, $G$ generates an image $x^g$ which can be passed to $T^s$ for better recognition results (as expressed in Assertion~\ref{eq:proba_accu}).
We oppose to $G$ a discriminator network \hbox{$D(x^a_z, x\ ; \theta_D)\to d$} defined by its hyper-parameters $\theta_D$ which, given a pair of images $(x^a_z, x)$ with $x^a_z = A(x^s, z)$, estimates the probability $d$ that $x$ is the original noiseless sample $x^s$ and not the recovered image $x^g = G(x^a_z)$~\cite{isola2016image}. 
% More formally, $d$ is thus an estimation of the probability $\Pr( x = x^g \mid x^a_z, x)$.

The typical objective such a solution has to maximize is:
\begin{align}
    G^* = {}& \arg \min_G \max_D \alpha \mathcal{L}_{d}(G, D) + \beta \mathcal{L}_{g}(G) \label{eq:minmax} \\
    \shortintertext{with}
    \mathcal{L}_{d}(G, D) = {}& \mathbb{E}_{x^s, z} \big[log D(x^a_z, x^s\ ; \theta_D)\big] + \\
& \mathbb{E}_{x^s} \Big[log \Big(1 - D\big(x^a_z, G(x^a_z\ ; \theta_G)\ ; \theta_D\big)\Big)\Big] \nonumber \\
\nonumber \\[-2ex]
	\mathcal{L}_{g}(G) = {} & \mathbb{E}_{x^s, z} \Big[\big\|x^s - G(x^a_z\ ; \theta_G)\big\|_1\Big]
\end{align}

As explained in~\cite{isola2016image,shrivastava2016learning,bousmalis2016unsupervised}, the conditional loss function $\mathcal{L}_{d}$, weighted by a coefficient $\alpha$ represents the cross-entropy error for a classification problem where $D(x^a_z, x\ ; \theta_D)$ estimates if $(x^a_z, x)$ is a ``fake'' or ``real'' pair (\ie $x$ generated by $G$ from $x^a_z$, or $x$ original sample from $X^s$).

This is complemented by a simple similarity loss $\mathcal{L}_{g}$, an $L1$ distance weighted by a parameter $\beta$, to force the generator to stay close to the ground-truth. 
However, since the images we are comparing are supposed to be noiseless with no background, this loss can be augmented with another one specifically targeting the foreground similarity (while we still want $\mathcal{L}_{g}$ to compare the whole images to make sure $G$ deals properly with backgrounds). 
We thus introduce a complementary \textit{foreground} loss $\mathcal{L}_{f}$ weighted by a factor $\gamma$, inspired by the \textit{content-similarity} loss of Bousmalis~\etal~\cite{bousmalis2016unsupervised}.
Given $m^s$ the binary foreground mask obtained from $x^s$ ($m^s_{ij} = 1$ if $x^s_{ij} \neq 0$ else $m^s_{ij} = 0$) and $\odot$ the Hadamard product:
\begin{equation}
\mathcal{L}_{f}(G) = \mathbb{E}_{x^s, z} \Big[
\big\|(x^s - G(x^a_z\ ; \theta_G))\odot m^s \big\|_1\Big] 
\end{equation}
In the case that the target recognition method $T^s(x) \to \widetilde{y}$ is provided ready-to-use for the GAN training, a third task-specific loss can be applied (similarly to~\cite{bousmalis2016unsupervised,li2017triple}, but with a fixed pre-trained network). Weighted by another parameter $\delta$, $\mathcal{L}_{t}$ can be used to guide $G$, to make it more \textit{aware} of the information this specific $T^s$ tries to uncover:
%\begin{equation}
%\mathcal{L}_{t} = \mathbb{E}_{x^s, y^s, z} \Big[\big\|y^s - T^s\big(G(x^a_z\ ; \theta_G)\big)\big\|_2\Big]
%\end{equation}
\begin{equation} \label{eq:task_specific_loss}
\mathcal{L}_{t}(G) = \mathbb{E}_{x^s, z} \Big[\big\|T^s(x^s) - T^s\big(G(x^a_z\ ; \theta_G)\big)\big\|_2\Big]
\end{equation}
This formulation has two advantages: no assumptions are made regarding the nature of $\widetilde{y} = T^s(x)$, and no ground-truth $y$ is needed since $\mathcal{L}_{t}$ only depends on the difference between the two estimations made by $T^s$. Unlike PixelDA~\cite{bousmalis2016unsupervised}, our training is thus both uncoupled from the real domain (no real image $x^r$ used) and completely unsupervised (no prior on the nature of the labels/features, and neither ground-truth $y^r$ of $x^r$ nor $y^s$ of $x^s$ used).
%By taking this optional loss $\mathcal{L}_{t}$ into account for the training of $G$, we push the generator not only to output the synthetic-looking equivalents of the input data, but also to preserve/enhance the information this particular $T^s$ is looking for.
Taking into account the newly introduced $\mathcal{L}_{f}$ and optional $\mathcal{L}_{t}$, the expanded loss guiding our method toward its objective is thus finally:
\begin{equation}
G^* = \alpha \mathcal{L}_{d}(G, D) + \beta \mathcal{L}_{g}(G)
 + \gamma \mathcal{L}_{f}(G) + \delta \mathcal{L}_{t}(G)
\end{equation}

Following the standard procedure~\cite{goodfellow2014generative}, the minmax optimization (with or without the term $\delta\mathcal{L}_{t}$) is achieved by alternating at each training iteration between (1) fixing $\theta_G$ to train $D$ over a batch of $(x^a_z, x^s)$ and  $\big(x^a_z, G(x^s\ ; \theta_G)\big)$ pairs, updating through gradient descent $\theta_D$ to maximize $\mathcal{L}_{d}$; (2) fixing $\theta_D$ to train $G$, updating through gradient descent $\theta_G$ to minimize the combined loss.

\subsection{Synthetic Data Generation} \label{sec:data_generation}

A key prior in this work is the unavailability of target real images (not only target labels). Only synthetic depth images $x^s$ are used to train our solution, and presumably the recognition method (if anecdotal target images $x^r$ are however available, they can obviously be used for fine-tuning).
The only requirement of our pipeline is thus the availability of the synthetic data used to train $T^s$, or at least the 3D models of the class objects $C$ to generate $X^s$ from them.

The first step of our solution thus consists of a depth image rendering pipeline, which offers several rendering strategies (\eg single-object rendering for classification or pose estimation applications, or random multi-object generation for detection) and viewpoint sampling methods (\eg user-provided lists, simulated turntable, \etc).
The most straightforward strategy used for our experiments replicates the method by Wohlhart~\etal~\cite{Wohlhart15,zakharov2017}: viewpoints are defined as vertices of an icosahedron centered on the target object(s). By repeatedly subdividing each triangular face of the icosahedron, additional viewpoints, and therefore a denser representation, can be created.
Furthermore, in-plane rotations can be added at each position by rotating the camera around the axis pointing at the object center.

Rotation invariances of the objects can also be taken into account (\eg for pose estimation applications), as samples of rotation-invariant objects representing different poses might look exactly the same, confusing the recognition methods. 
To deal with this, the pipeline can be configured to trim the number of poses for those objects, such that every image is unique; as done in ~\cite{kehl2017ssd,zakharov2017}. 

\subsection{Inline Data Augmentation} \label{sec:augmentation}

\begin{figure*}[t]
  \centering
  \includegraphics[width=1\linewidth]{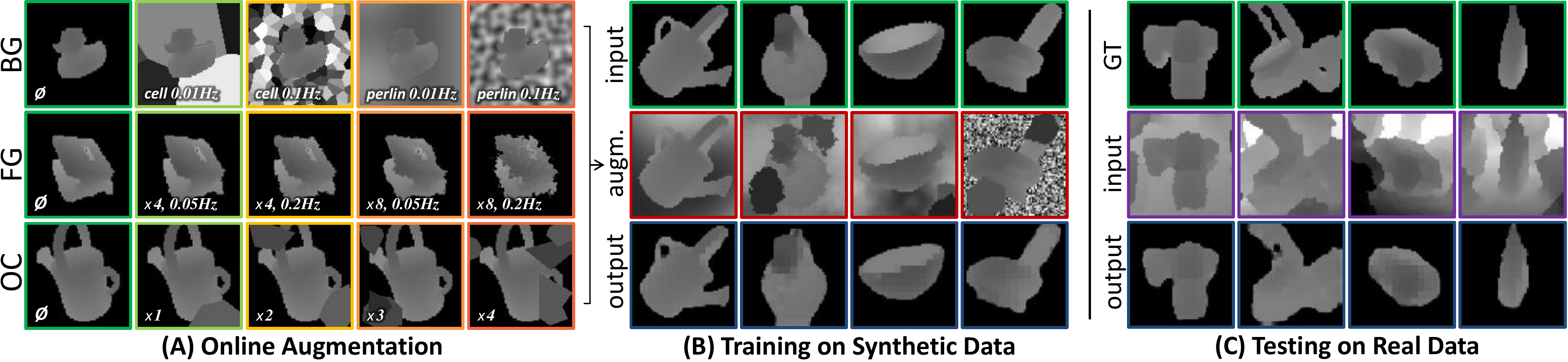}
  \caption{\textbf{(A) Augmentation examples} (\textit{BG} background noise; \textit{FG} foreground distortion; \textit{OC} occlusions) for different noise amplitudes or types; \textbf{(B) Validation results}, showing how our solution learns during training to recover the clean images (\textit{input}, here from LineMOD~\cite{hinterstoisser2012model}) from their augmented versions (\textit{augm.}); \textbf{(C) Test results on real data} (compared to ground-truth \textit{GT}).}
  \label{fig:augmentation_and_results}  
\end{figure*}

With no target images to teach the generator $G$ how to map the real images to their synthetic (noiseless and segmented) equivalents, we make $G$ play against $A(x^s, z) \to x^a_z$. 
$A$ is an extensive inline augmentation pipeline which applies a series of transformations (as shown in Figure~\ref{fig:augmentation_and_results}) parametrized by a vector $z$ randomly sampled for every image in the training batches, at every iteration.

The goal of $A$ is \textit{not} to generate realistic images out of $X^s$, \ie not to obtain $X^a=\Set{A(x^s_{i \bmod |X^s|}, z_i)}{\forall i \in N^{iter}} \approx X^r$, but instead to reach $X^r \subsetneq X^a$ approximately true for a large enough number of training iterations $N^{iter}$ and a large enough set of stochastic transformations applied.
Since the vector $z$ is sampled from a $k$-dimensional finite set $\mathbb{Z}^k = \{0,...,n_i\}_{i=0}^k$ with $k$ number of augmentation parameters and $n_i$ their maximum values;
the probabilities of random transformations to be applied, and their amplitudes, are directly linked to $|\mathbb{Z}^k| = \sum_{i=0}^{k} (n_i+1)$.
Leaving $N^{iter}$ aside, the selection of $n_0,...,n_k$  can thus be seen as a higher-level minmax procedure between the GAN and the augmentation process. The GAN is trained to optimize $G^*$ while $A$ tries to prevent it.
This led to a series of meta-iterations to make $A$ challenging enough (\ie to fix $\mathbb{Z}^k$ as large as possible without tampering images beyond reversibility) so that $G$ can learn robust features to discriminate and recover the noiseless signal.

Inspired by the literature~\cite{simard2003best,ciregan2012multi,chatfield2014return}, the procedures composing $A$ are the following:

\par\noindent
\textbf{Sensor noise and 3D clutter:}
With the class and pose for every image $x^s$ easily available from the rendering engine, a state-of-the-art simulation pipeline, \eg \textit{DepthSynth} provided by Planche \etal~\cite{planche2017depthsynth}, can be used to generate a pseudo-realistic depth image to replace the noiseless $x^s$. 
This simulation reproduces the mechanisms of structured-light sensors to achieve similar noise (\eg shadow noise, missing data from improper pattern reflection, mismatching during the stereo-matching process, \etc), and can also be used to clutter the images by directly adding 3D elements into the rendered scene (\eg ground floor, random shapes, \etc). 
This single augmentation step could be enough if the simulation was indeed achieving proper realism. Experiments however showed that it can be heavily affected by the quality of the 3D models and by their actual lack of proper reflectance model. Moreover, rendering cluttered 3D scenes is computationally expensive.
As a result, only a random subset of $X^s$ undergoes this process every iteration (the size of the subset being defined by one of the random variables of $z$); while the following two-dimensional transformations do most of the augmentation, and partially compensate for the biases of the simulation pipeline.

%\par\noindent
%\textbf{Linear transforms:}
%The content of the training images may undergo small 2D translations \eg to cover cases when detected objects haven't been properly centered in their patches at detection time. 
%However, one doesn’t want to apply too large linear transforms, or $G$ may start taking into account peripheral elements when processing the cluttered data (\eg other target objects appearing in the background).
%\cBen{rewrite more formally / remove since not used?}

\par\noindent
\textbf{Background noise:}
To teach $G$ to focus on the representations of the captured objects and ignore the rest, $A$ fills the background of the training data with several noise types commonly used in procedural content generation \eg fractal Perlin noise~\cite{perlin2002improving}, Voronoi texturing~\cite{worley1996cellular}, and white noise. These  patterns are computed using a vast frequency range (function of $z$), further increasing the number of possible background variations.  

\par\noindent
\textbf{Foreground distortion:} 
Similarly to Simard~\etal~\cite{simard2003best}, images undergo random distortions (as an inexpensive way to simulate sensor noise, objects wear-and-tear, \etc).
A 3D vector field is generated using the Perlin noise method~\cite{perlin2002improving} and applied as offset values to each image pixel, causing the warping. Since noise values range from $-1$ to $1$  by design, we introduce a multiplicative factor, which allows for more severe distortions.

\par\noindent
\textbf{Random occlusions:}
Occlusions are introduced to serve two different purposes: to teach $G$ to reconstruct the parts of the objects which may be partially occluded; 
and to further enforce invariance within the depth scans to additional objects which don't belong to the target classes $C$ (\ie to ignore them, treat them as part of the background).
Based on~\cite{ounsworth2015anticipatory}, occlusion elements are generated by walking around the circle taking random angular steps and random radii at each step. Then the generated polygons are filled with arbitrary depth values and painted on top of the patch.

\subsection{Network Architectures}
\label{sec:archi}
%\begin{figure}[t]
%  \centering
%  \includegraphics[width=1\linewidth]{figures/generator-architecture}
%  \caption{\textbf{Architecture of the generator network} \cBen{UPDATE WITH FINAL SUPERPARAMETERS OR REMOVE}}
%  \label{fig:archi-gen} 
%\end{figure}
%
%\begin{figure}[t]
%  \centering
%  \includegraphics[width=1\linewidth]{figures/discriminator-architecture}
%  \caption{\textbf{Architecture of the discriminator network} \cBen{UPDATE WITH FINAL SUPERPARAMETERS OR REMOVE}}
%  \label{fig:archi-disc} 
%\end{figure}

Though our \textit{UnrealDA} is not bound to particular network architectures, we opted for the work of Isola~\etal~\cite{isola2016image}, chosen for its efficiency and popularity at the time of this paper.
The generator $G$ is a U-Net~\cite{ronneberger2015u} with skip connections between each encoding block and its opposite decoding one. 
Building on Isola~\etal formalism~\cite{isola2016image}, each encoding block $B^C_k$ is made of the following layers: 2-factor downsampling Convolution with $k$ filters, BatchNorm (except for $1^{st}$ block), then LeakyReLU. 
Each decoding block $B^D_k$ is made of: 2-factor upsampling Transposed Convolution with $k$ filters, BatchNorm, LeakyReLU, then $\frac{1}{2}$-rate Dropout layers. All Convolution layers have $4 \times 4$ filters, and Dropout layers have a $0.2$ slope. The generator thus consists of: $B^C_{64}-B^C_{128}-B^C_{256}-B^C_{512}-B^C_{512}-B^C_{512}-B^C_{512}-B^C_{512}-B^D_{512}-B^D_{1024}-B^D_{1024}-B^D_{1024}-B^D_{1024}-B^D_{512}-B^D_{256}-B^D_{128}$.
Similarly, the discriminator follows the $70 \times 70$ \textit{PatchGAN} architecture~\cite{isola2016image}:  $B^C_{64}-B^C_{128}-B^C_{256}-B^C_{512}$. 
Further implementation details (for the networks and also the augmentations) can be found in the supplementary material.
%
%\subsection{Fine Tuning}
%If available, real depth scans can be used to fine-tune the method. For each real image, a 3D model of its foreground and the viewpoint information is needed as ground-truth.
%Using the 3D engine configured to generate noiseless depth images, clean images of the foreground from the same viewpoints can thus be generated. Each of these synthetic images are used both:
%\begin{itemize}
%\item as a mask to crop the foreground out of the real image, obtaining a background-less real scan which will be used as target of the first GAN and input of the second GAN;
%\item as the target image of the second GAN.
%\end{itemize}
%
%
%\subsection{Usage}
%
%Once trained, the proposed pipeline can simply be used on every real depth scans containing one of the target objects, to extract and clean its depth information. The result can then be used for various applications \eg instance recognition or pose estimation.

%-------------------------------------------------------------------------
\section{Evaluation}
\label{sec:exp}
% !TeX spellcheck = en_US

\begin{table*}[t]
%\footnotesize
	\centering
	\caption{
\textbf{Quantitative results on different tasks and datasets}: (A) Classification accuracy of different instances of a simple LeNet network $T_{le}$~\cite{Lecun98} over a subset of T-LESS (5 objects); (B) Classification and angular accuracy of different instances of the triplet method $T_{tri}$~\cite{zakharov2017} over LineMOD (15 objects). Instances were trained on various data modalities (noiseless synthetic for $T^s$; synthetic augmented for $T^a$; or real for $T^r$) and tested on the real datasets $X_{test}^r$ with different pre-processing (none; pre-processing by $G^a$ purely trained on synthetic augmented data; or by $G^r$ same method trained over a mix of real and augmented data). For each line, the angular accuracy is computed only on the subset of properly classified data.
}

	\label{tab:icpe_and_lenet} 
	\resizebox{1\linewidth}{!}{
		\def\arraystretch{1}
		\begin{tabu}{@{}cc@{\hskip 10pt}c@{\hskip 5pt}|@{\hskip 5pt}cc@{\hskip 010pt}c@{\hskip 15pt}cc@{}}
\multicolumn{3}{c}{(A) IC on T-LESS ($T_{le} =$ LeNet)} 					& 	
\multicolumn{5}{c}{(B) ICPE on LineMOD  ($T_{tri} =$ Triplet Method)}		\\
\toprule

\multirow{2}{*}{\textbf{Input}} & \multirow{2}{*}{\textbf{Method}} & \multirow{2}{*}{\textbf{\shortstack{Classification\\accuracy}}}	&
\multirow{2}{*}{\textbf{Input}} & \multirow{2}{*}{\textbf{Method}} & \multicolumn{2}{c}{\textbf{Angular accuracy}} & {\hskip 5pt}\multirow{2}{*}{\shortstack{\textbf{Classification}\\\textbf{accuracy}}}
			\\ \cline{6-7}
& & & & & \rule[5pt]{0pt}{5pt} \textbf{Median} & \textbf{Mean} \\
			\midrule
			%\cmidrule(r){1-1} \cmidrule(l){5-5} 
			$\pmb{X_{test}^r}$ & $\pmb{T^s_{le}}$ & 20.48\% & $\pmb{X_{test}^r}$ & $\pmb{T^s_{tri}}$ & 93.48$^{\circ}$ &  100.28$^{\circ}$  & {\hskip 5pt} 7.71\% \\
			$\pmb{X_{test}^r}$ & $\pmb{T^a_{le}}$ & 83.35\% & $\pmb{X_{test}^r}$ & $\pmb{T^a_{tri}}$ & 13.45$^{\circ}$ &  30.07$^{\circ}$ & {\hskip 5pt} 82.14\% \\
			\rowfont{\color{blue}}
			$\pmb{G^a(X_{test}^r)}$ & $\pmb{T^s_{le}}$ & 93.01\% & $\pmb{G^a(X_{test}^r)}$ & $\pmb{T^s_{tri}}$ & 13.74$^{\circ}$ & 31.14$^{\circ}$ & {\hskip 5pt} 94.77\% \\
			\midrule
			$\pmb{X_{test}^r}$ & $\pmb{T^r_{le}}$ & 95.92\% & $\pmb{X_{test}^r}$ & $\pmb{T^r_{tri}}$ & 12.13$^{\circ}$ & 27.80$^{\circ}$ & {\hskip 5pt} 95.49\% \\ 
			\rowfont{\color{blueviolet}}
			$\pmb{G^r(X_{test}^r)}$ & $\pmb{T^s_{le}}$ & 96.67\% & $\pmb{G^r(X_{test}^r)}$ & $\pmb{T^s_{tri}}$ & 11.64$^{\circ}$ &  24.31$^{\circ}$ & {\hskip 5pt} 98.44\% \\ 
			\bottomrule
		\end{tabu}
	%\vspace{-0.4cm}   
	}
\end{table*}

In this section, we perform a thorough analysis to demonstrate the effectiveness of our \textit{UnrealDA} pipeline and its components over a variety of datasets and tasks, and to compare against state-of-the-art domain adaptation methods.

\subsection{Datasets} \label{sec:datasets}

\par\noindent
\textbf{T-LESS}: T-LESS~\cite{hodan2017t} is a challenging RGB-D dataset with 3D models for detection, containing industrial objects of similar shapes, often heavily occluded. For a preliminary experiment, we consider the first three scenes captured with a Primesense Carmine sensor, building a subset of {\raise.17ex\hbox{$\scriptstyle\mathtt{\sim}$}}2.000 depth patches from 5 objects, occluded up to 60\%.% (numbers 2, 6, 7, 25, 29)

\par\noindent
\textbf{LineMOD}:
The LineMOD dataset~\cite{hinterstoisser2012model} has been chosen as the main evaluation dataset. It contains 15 mesh models of distinctive objects and their RGB-D sequences together with camera poses. This dataset has four symmetric objects (\textit{cup}, \textit{bowl}, glue, and \textit{eggbox}), which result in ambiguities for the task-specific pose estimation algorithm (similar looking patches might have completely different poses). To resolve this problem, we constrain real views by keeping only unambiguous poses for these four objects.

\par\noindent
\textbf{Data preparation}: Synthetic images are generated following the exemplary procedure described in Section~\ref{sec:data_generation}, using a simple 3D engine to generate \textit{z-buffer} scans from 3D models, obtaining patches centered on the objects of interest. Since the test datasets contain in-plane rotations, we generate the training samples taking this additional degree of freedom into account.
For each dataset, the real images are split in two subsets: 50\% of them compose $X^r_{test}$, the test set; while the other 50\% ($X^r_{train}$) are used to train methods our synthetic-only pipeline is set to compete against.

\subsection{Task-Specific Algorithms}

For quantitative evaluation, we integrate our pipeline to different recognition methods, measuring the impact.
We first consider instance classification (IC) with a simple method $T_{le}$. Based on the LeNet architecture~\cite{Lecun98}, this network takes a depth image as input and returns its estimated class in the form of a softmax probability layer. A cross-entropy loss is used for its training.
We define as $T^s_{le}$ the algorithm purely trained on synthetic samples from $X^s$ and their labels.

Besides this simple classifier, we consider a more complex recognition method to demonstrate that our solution is not tailored to a specific pipeline.
We thus choose a method for instance classification and pose estimation (ICPE) closely following the implementation of \cite{zakharov2017}. This algorithm $T_{tri}$ uses a so-called triplet CNN to map image patches to a lower-dimensional descriptor space, where object instances and their poses are well separated. To be able to learn this mapping, the network weights are adjusted to minimize the following loss:
\begin{align}
&\mathcal{L}_{tri} = \mkern-22mu
\sum_{(x_b,x_p,x_n) \in X} { \mkern-22mu max\left(0,1-\frac{||T_{tri}(x_b)-T_{tri}(x_n)||_2^2}{||T_{tri}(x_b)-T_{tri}(x_p)||_2^2+m}\right)} \nonumber \\
&\text{where } m =
\begin{cases}
2 \arccos (|q_b \cdot q_p|) &\text{if } c_b = c_p,\\
n &\text{else, for } n > \pi.
\end{cases}
\end{align}
with $x_b$ the input image used as binding anchor , $x_p$ a positive or similar sample, and $x_n$ a negative or dissimilar one. $T_{tri}(x)$ is the feature returned by the network given the image $x$, and $m$ is the margin setting the minimum ratio for the distance between similar and dissimilar pairs of samples.

Once trained, the network is used to compute the features of a subset of $X^s$, stored together with object instances and poses to form a feature-descriptor dataset $X_{db}^s$. Recognition is done on test data by using the trained network to compute the descriptor of each provided image ($x^r$ or $G(x^r)$) then applying a nearest neighbor search algorithm to find its closest descriptor in $X_{db}^s$.

%
%\cBen{"Implementation Details" subsection here? \eg Adam optimizer with such parameters, values of the weights $\alpha$, $\beta$, \etc Even move the Architectures here?}

\subsection{Experiments and Discussions}

\begin{table*}[t]
	\centering
	\caption{
	\textbf{Comparison to opposite domain-adaptation GANs}: given the two recognition tasks ``(A) Instance Classification on T-lESS (5 objects) with $T_{le}$" and ``(B) Instance Classification and Pose Estimation on LineMOD (15 objects) with $T_{tri}$" (defined for the experiment in Table~\ref{tab:icpe_and_lenet}), we train several modalities of the networks $T$ against diverse domain-adaptation GANs trained on real data $X_{train}^r$ ($50\%$ of the datasets), and compare their final accuracy with our results.}
	\label{tab:gans_exp} 
	\resizebox{1\linewidth}{!}{
		\def\arraystretch{1}
		\begin{tabu}{@{}c@{\hskip 10pt}c@{\hskip 10pt}cc@{\hskip 8pt}|c|@{\hskip 10pt}c@{\hskip 10pt}cc@{}}
\multicolumn{4}{c}{}  & \multicolumn{1}{c}{(A) IC on T-LESS with $T_{le}$}{\hskip 10pt} & \multicolumn{3}{c}{(B) ICPE on LineMOD  with $T_{tri}$}		\\

\toprule

& \multicolumn{2}{c}{\textbf{$\pmb{T}$ Modality}} & & \multirow{2}{*}{\textbf{\shortstack{Classification\\accuracy}}}	&

 \multicolumn{2}{c}{\textbf{Angular accuracy}} & \multirow{2}{*}{{\hskip 10pt}\shortstack{\textbf{Classification}\\\textbf{accuracy}}}
			\\ \cline{2-3} \cline{6-7}
 & \textbf{Trained on} & \textbf{Applied to} & &  & \rule[5pt]{0pt}{5pt} \textbf{Median} & \textbf{Mean}  & \\
			\midrule

\parbox[t]{6mm}{\multirow{4}{*}{\rotatebox[origin=c]{90}{\textbf{\shortstack{Requiring\\Real Data}}}}}
& \textbf{\textit{CycleGAN}} & $\pmb{X_{test}^r}$ & & 40.97\% & 71.10$^{\circ}$ & 86.73$^{\circ}$ & {\hskip 10pt}14.72\% \\
& \textbf{\textit{SimGAN}} & $\pmb{X_{test}^r}$ & & 59.20\% & 20.44$^{\circ}$ & 43.36$^{\circ}$ & {\hskip 10pt}73.20\% \\
& \textbf{\textit{PixelDA}} & $\pmb{X_{test}^r}$ & & 89.75\% & 18.15$^{\circ}$ & 39.06$^{\circ}$ & {\hskip 10pt}90.31\% \\
\rowfont{\color{blueviolet}}
& \textbf{\textit{$\pmb{X^s}$}} & $\pmb{G^r(X_{test}^r)}$ & & 96.67\% & 11.64$^{\circ}$ &  24.31$^{\circ}$ & {\hskip 10pt} 98.44\% \\
			\midrule
\rowfont{\color{blue}}
& \textbf{\textit{$\pmb{X^s}$}} & $\pmb{G^a(X_{test}^r)}$ & & 93.01\% & 13.74$^{\circ}$ & 31.14$^{\circ}$ & {\hskip 10pt} 94.77\% \\
			
			\bottomrule
		\end{tabu}
	}
%	\vspace{0.4cm}   
\end{table*}

\subsubsection{Comparisons with Different Baselines}

We first demonstrate the effectiveness of our data pre-processing on a set of different tasks, and show the benefits of decoupling this operation from recognition itself.
As a preliminary experiment, we define an instance classification task on the T-LESS patch dataset, with the LeNet network.

One could argue that directly training $T_{le}$ on augmented synthetic data could be more straight-forward than training $G$ against $A$ and plugging it in over $T^s_{le}$ afterwards.
To prove that our solution not only has the advantage of uncoupling the training of recognition methods to the data augmentation but also improves the end accuracy, we introduce $T^a_{le}$ the algorithm trained on augmented data from $A$. 
We additionally define method $T^r_{le}$ trained on 50\% of the real data.
During test time, the real depth patches are either directly handed to the classifiers, pre-processed by our generator $G$ (exclusively trained on synthetic data, renamed $G^a$ here for clarity), or pre-processed by a generator $G^r$. Using the same pipeline, $G^r$ is trained on a combination of augmented synthetic data and real data ($X^r_{train}$); and thus serves as a theoretical upper performance bound for the GANs.
Evaluation is done for different combinations of pre-processing and recognition modalities, computing the final classification accuracy for each. Results are shown in Table~\ref{tab:icpe_and_lenet}-A.
To demonstrate how \textit{UnrealDA} generalizes both to different datasets (with diverse classes and environments) and to distinctive task-specific applications, we reproduce the same experimental protocol on a more complex ICPE task. 
Following the previous notations, we define as $T^s_{tri}$ the triplet method trained on synthetic data $X^s$; $T^a_{tri}$ the same algorithm trained on augmented one; and $T^r_{tri}$ trained on real data $X^r_{train}$.
These different instances are used along their respective feature-descriptor dataset $X_{db}^s$ to classify instances from LineMOD (15 objects) and estimate their 3D poses. Once again, the networks are either handed unprocessed test data, data pre-processed by $G^a$, or data pre-processed by $G^r$. If the class of each returned descriptor agrees with the ground-truth, we compute the angular error between their poses. Once this procedure performed on the entire set $X^r_{test}$, the classification and angular accuracy (as median and mean angles) are used for the comparison shown Table~\ref{tab:icpe_and_lenet}-B.

For both experiments, we consistently observe the positive impact \textit{UnrealDA}'s pre-processing has on the recognition. The performance of the algorithms using our modality $G^a$ (trained exclusively on synthetic data) matches the results of those plugged to $G^r$, trained on real data. This demonstrates the effectiveness of our advanced depth data augmentation pipeline. Improvements could be done to match this higher-bound baseline by tailoring the augmentation pipeline to a specific sensor or environment. Our current augmentation pipeline has however the advantage of genericity. While $G^r$ is only trained for the sensor and background type(s) of the provided real dataset, our solution $G^a$ has been trained over $A$ for domain invariance.

\begin{table*}[t]
	\centering
	\caption{
	\textbf{Quantitative ablation study}: given the two recognition tasks ``(A) Instance Classification on T-lESS (5 objects) with $T_{le}$" and ``(B) Instance Classification and Pose Estimation on LineMOD (15 objects) with $T_{tri}$" (defined for the experiment in Table~\ref{tab:icpe_and_lenet}), we train both networks on noiseless data and evaluate them on the outputs of different modalities of $G^a$. Each is either trained (A) with different augmentation combinations (\textit{BG} background noise; \textit{FG} foreground distortion; \textit{OC} occlusions; \textit{SI} sensor simulation); or (B) with different loss combinations ($\mathcal{L}_{d}+\mathcal{L}_{g}$ vanilla GAN loss; $\mathcal{L}_{f}$ foreground-similarity loss; $\mathcal{L}_{t}$ task-specific loss).}
	\label{tab:aug_and_losses} 
	\resizebox{1\linewidth}{!}{
		\def\arraystretch{1}
		\begin{tabular}{@{}c@{\hskip 10pt}l@{\hskip 10pt}|>{\centering\arraybackslash}p{3.8cm}|@{\hskip 10pt}c@{\hskip 10pt}cc@{}}
\multicolumn{2}{c}{}  & \multicolumn{1}{c}{(A) IC on T-LESS with $T^s_{le}$}{\hskip 10pt} & \multicolumn{3}{c}{(B) ICPE on LineMOD  with $T^s_{tri}$}		\\

\toprule

& \multirow{2}{*}{\textbf{Modality}} & \multirow{2}{*}{\textbf{\shortstack{Classification\\accuracy}}}	&

 \multicolumn{2}{c}{\textbf{Angular accuracy}} & \multirow{2}{*}{{\hskip 10pt}\shortstack{\textbf{Classification}\\\textbf{accuracy}}}
			\\ \cline{4-5}
 & & & \rule[5pt]{0pt}{5pt} \textbf{Median} & \textbf{Mean}  & \\
			\midrule

\parbox[t]{6mm}{\multirow{4}{*}{\rotatebox[origin=c]{90}{\textbf{\shortstack{(i)\\Augment.}}}}}
& \textbf{\textit{BG}} & 79.93\% & 17.64$^{\circ}$ & 38.90$^{\circ}$ & {\hskip 10pt} 83.78\% \\
			& \textbf{\textit{BG+SI}} & 84.09\% & 17.60$^{\circ}$ & 42.22$^{\circ}$ & {\hskip 10pt} 85.26\% \\
			& \textbf{\textit{BG+FG}} & 89.42\% & 15.25$^{\circ}$ & 34.90$^{\circ}$ & {\hskip 10pt} 92.33\% \\

			& \textbf{\textit{BG+FG+OC}} & 93.01\% & 13.74$^{\circ}$ & 31.14$^{\circ}$ & {\hskip 10pt} 94.77\% \\
			\midrule
			%\cmidrule(r){1-1} \cmidrule(l){5-5} 
			%			\textbf{\textit{None}}  & 0.02\% &  0.16\% & 1.72\% & 11.08\% \\
			\parbox[t]{6mm}{\multirow{3}{*}{\rotatebox[origin=c]{90}{\textbf{\shortstack{(ii)\\Losses}}}}}
			& $\pmb{\mathcal{L}_{d}+\mathcal{L}_{g}}$ & 91.09\% & 14.49$^{\circ}$ &  33.91$^{\circ}$ & {\hskip 10pt} 92.89\% \\
			& $\pmb{\mathcal{L}_{d}+\mathcal{L}_{g}+\mathcal{L}_{f}}$ & 92.17\% & 14.34$^{\circ}$ & 32.21$^{\circ}$ & {\hskip 10pt} 93.39\% \\
			& $\pmb{\mathcal{L}_{d}+\mathcal{L}_{g}+\mathcal{L}_{f}+\mathcal{L}_{t}}$ & 93.01\% & 13.74$^{\circ}$ & 31.14$^{\circ}$ & {\hskip 10pt} 94.77\% \\
			\bottomrule
		\end{tabular}
	}
%	\vspace{0.4cm}   
\end{table*}

We extended this study by also performing the ICPE on the real LineMOD scans with their backgrounds perfectly removed. The accuracy using $T^s_{tri}$ trained on pure synthetic data was only 67.32\% for classification, and 24.56$^{\circ}$ median / 51.58$^{\circ}$ mean for the pose estimation. This is well below the results on images processed by \textit{UnrealDA}; attesting that our pipeline not only does background subtraction but also effectively uses the CADs prior to recover clean geometry from noisy scans, improving recognition.

Finally, Table~\ref{tab:icpe_and_lenet} also reveals the accuracy improvements obtained by decoupling data augmentation and recognition training. 
Both recognition methods $T^s_{le}$ and $T^s_{tri}$ trained on ``pure'' noise-free $X^s$ performs better when used on top of our solution, compared to their respective architectures $T^a$ directly trained on augmented data. It is even comparable to their respective $T^r$ trained on real images.
This confirms our initial intuition regarding the advantages of teaching recognition methods in a noise-free controlled environment, to then map the real data into this known domain. 
Besides, this decoupling allows once again for greater reusability. Augmentation needs to be done only once to train $G$, which can then be part of any number of recognition pipelines.

\subsubsection{Comparison to Usual Domain Adaptation GANs}

As previous GAN-based methods to bridge the realism gap are using a subset of real data to learn a mapping from synthetic to realistic, it seems difficult to present a fair comparison to our opposite solution, trained on synthetic data only. We opted for a practical study on the aforementioned tasks, considering the end results given the same recognition methods and test sets. Selecting prominent solutions, SimGAN~\cite{shrivastava2016learning}, CycleGAN~\cite{zhu2017unpaired} and PixelDA~\cite{bousmalis2016unsupervised}, we trained them on $X^s$ and $X^r_{train}$ ($50\%$ of the real datasets) so they learn to generate pseudo-realistic images to train the methods $T$ on. For each task, we measure the modalities' accuracy on $X^r_{test}$, comparing with $T^s$ applied to $G^a(X_{test}^r)$ and $G^r(X_{test}^r)$.

As SimGAN is designed to refine the pre-existing content of images and not to generate new elements like backgrounds and occlusions, we help this method by filling the images with random background noise beforehand. Still, the refiner doesn't seem able to deal with the lack of concrete information and fails to converge properly.
Unlike the other candidates, CycleGAN neither constrains the original foreground appearance nor tries to regress semantic information to improve the adaptation. Even though the resulting images are filled with some pseudo-realistic clutter, the target objects are often distorted beyond recognition, impairing the task networks training.
Finally, PixelDA results look more realistic while preserving most of the semantic information, out of the box. This is achieved by training $T^s_{le}$ along its GAN. However, this training procedure is not directly compatible to some recognition architectures like $T^s_{tri}$ (as it requires a specific batch-generation process), which is thus trained afterwards on the adapted data.

These observations are confirmed by the results presented in Table~\ref{tab:gans_exp}, which attest of the effectiveness of our reverse processing  compared to state-of-the-art methods for these challenging tasks (similar-looking manufactured objects, occlusions, clutter, \etc).

\subsubsection{Evaluation of the Solution Components}

Performing an ablation study, we first demonstrate the importance of each component of the depth data augmentation pipeline on the final results. Using the same two tasks (``IC on T-LESS" and ``ICPE on LineMOD"), we train five different \textit{UnrealDA} instances using various degrees of data augmentation. %: background, foreground, occlusions, and sensor simulation. 
The results are shown in Table~\ref{tab:aug_and_losses}-A, with a steady increase in recognition accuracy for each augmentation component added to the training pipeline. 
This experiment also justifies our choice to replace sensor noise simulation by foreground distortion. Given the current state-of-the-art in 2.5D simulation (heavily affected by the quality of the 3D models themselves and often too deterministic), our warping solution is both lighter and more stochastic, and thus a better challenge to prepare $G$.

\begin{figure*}[t]
  \centering
  \includegraphics[width=1\linewidth]{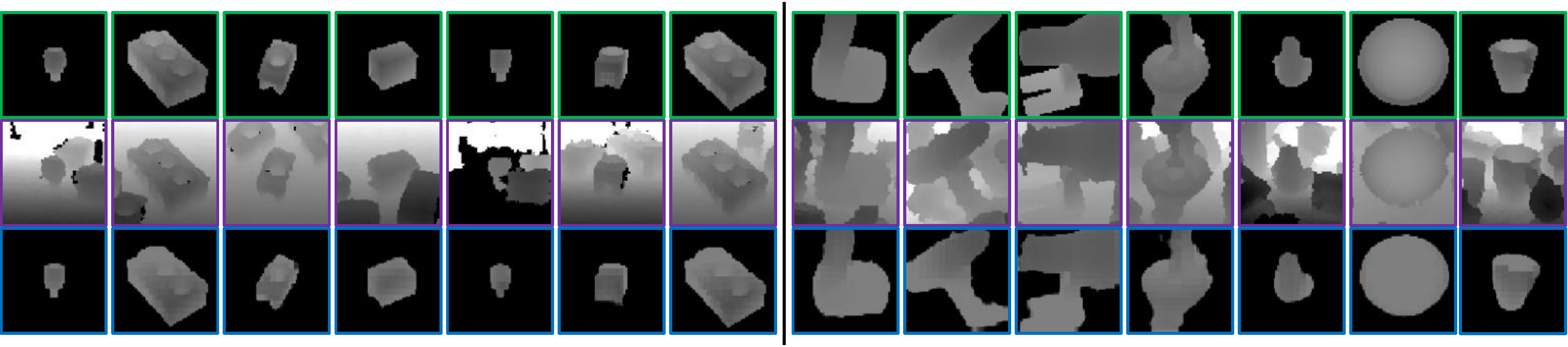}
  \caption{\textbf{Further qualitative results on T-LESS~\cite{hodan2017t} and LineMOD~\cite{hinterstoisser2012model}}. Processing of real scans (\textit{input}) by our purely-synthetic method, compared to the rendered ground-truth (\textit{GT}). More results can be found in the supplementary material}
  \label{fig:tless}  
\end{figure*}
We set up another experiment to evaluate the influence of the supplementary losses on the end results, as displayed in Table~\ref{tab:aug_and_losses}-B.
%\sout{As mentioned in Section~\ref{sec:learning_process}, $\mathcal{L}_{f}$ puts more attention to the image foreground, \ie object of interest, trying to better preserve its shape and content. 
%The task-specific loss $\mathcal{L}_{t}$, on the other hand, is optional and can be used to tailor $G$ to a provided $T^s$, forcing $G$ to preserve or recover the features $T^s$ was trained to look for. Each loss addition brings our purely-synthetic pipeline consistently closer to the higher bound results obtained by training over a large subset of target real data and their labels. }
It shows for instance that the classification \textit{error rate} on LineMOD drops from 7.11\% with the vanilla losses to 5.23\% with $\mathcal{L}_f$ and $\mathcal{L}_t$, \ie a 26\% relative drop which is rather significant, given the upper-bound error rates, \cf Table~\ref{tab:icpe_and_lenet}. If our augmentation pipeline and vanilla GAN already achieve great results, $\mathcal{L}_f$ and $\mathcal{L}_t$ can further improve them, depending on the sensor type for $\mathcal{L}_f$ or the specific tasks for $\mathcal{L}_t$, and so covering a wider range of sensors and applications.

Leveraging these components, our pipeline not only learns purely from synthetic data (and thus in a completely unsupervised manner) how to qualitatively denoise and declutter depth scans (\cf qualitative results in Figure~\ref{fig:tless}), but also considerably improves the performance of recognition methods using it to pre-process their input. We demonstrated how our solution makes such algorithms, simply trained on rendered images, almost on a par with the same methods trained in a supervised manner, on images from the real target domain.

%-------------------------------------------------------------------------
\section{Conclusion}
\label{sec:cnc}
% !TeX spellcheck = en_US

We presented an end-to-end pipeline requiring only the 3D models of the target objects to learn how to pre-process their real depth scans. Without accessing any real data or constraining the recognition methods during training, our solution tackles the {\em realism gap} problem in a simple, yet novel and effective manner.
Adapting a GAN architecture and relying on an extensive data augmentation process, our pipeline not only generalizes to many tasks by decoupling recognition and domain invariance, but also improves their end results.
We believe this concept will prove itself greatly useful to the community, and intend to apply it to other applications in the future.

%We thus believe this concept will prove itself greatly useful to the community, leveraging the parallel efforts to gather detailed 3D datasets, to develop more advanced simulation tools, and to improve GAN-based processing solutions, in order to bridge the data discrepancy gap from both sides.

%\cBen{ADD FUTURE WORK? \eg extending to RGB modalities despite the difficulty given the usual lack of texture}

% Despite the maturity of our proposed concept, we are well aware---through the previous results and our own advanced use cases---it has room for improvement, such as more sophisticated background modeling, occlusion handling and enhanced reflectance acquisition (\eg as a Bidirectional Reflectance Distribution Function (BRDF)).

\newpage

%-------------------------------------------------------------------------
\appendix
\beginsupplement

\noindent
{\Large\textbf{Supplementary Material}}

% !TeX spellcheck = en_US
\section{Schematic Overview of the Different Gap-Bridging Methods}

Table~\ref{tab:gan_comp} contains a schematic comparison of the training and testing solutions for recognition tasks addressed in the paper, when real images from the target domain(s) are available for training or not. 

Our \textit{UnrealDA} is the only approach focusing on pre-processing the real test scans instead of the synthetic training dataset, and on leveraging augmentation operations to perform well when no real data are available for training. Furthermore, as the task-specific networks can be trained separately on pure synthetic data (not augmented or processed by another GAN), our solution can directly be applied to various models, even when already trained.

\begin{table*}
  \caption{\textbf{Visual comparison of training and testing schemes.}} \label{tab:gan_comp}
  \begin{tabular}{lc@{\hskip 5pt}|@{\hskip 5pt}l@{\hskip 10pt}|@{\hskip 5pt}r}
    & & Training & Testing\\
  \toprule
  
  \parbox[c]{6mm}{\multirow{17}{*}{\rotatebox[origin=c]{90}{\textbf{\color{red}{Real Data Available for Training}}}}}
  
      & \parbox[c]{6mm}{\rotatebox[origin=c]{90}{\shortstack{Naive \\ Approach}}} & 
      \parbox[c]{14em}{
      \includegraphics[width=\linewidth]{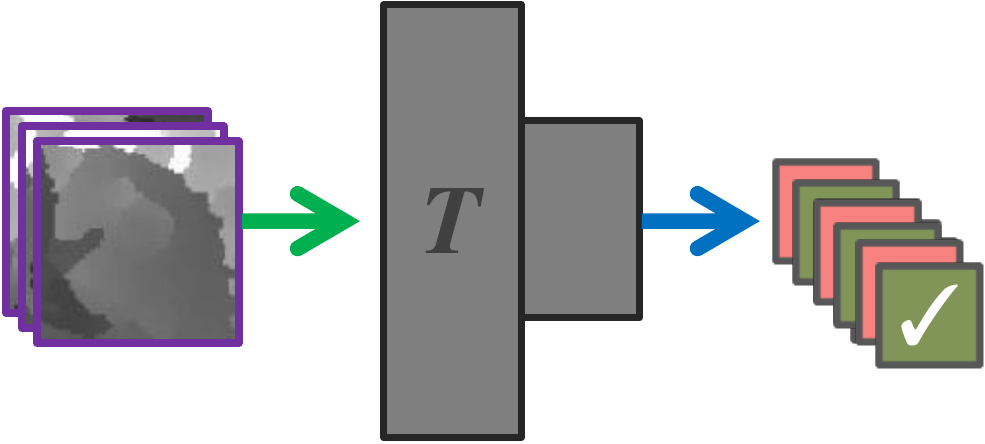}
      }
      &
      \parbox[c]{14em}{
      \includegraphics[width=\linewidth]{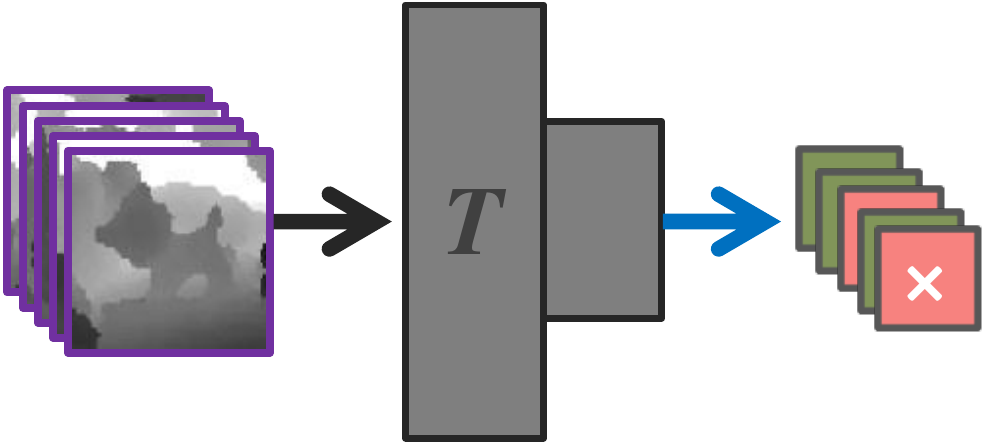}
      }
      \\[3.5em]
      
      & \parbox[c]{6mm}{\rotatebox[origin=c]{90}{\shortstack{Previous GAN \\ Approaches  \cite{shrivastava2016learning,bousmalis2016domain,zhu2017unpaired}}}} & 
      \parbox[c]{28em}{
      \includegraphics[width=\linewidth]{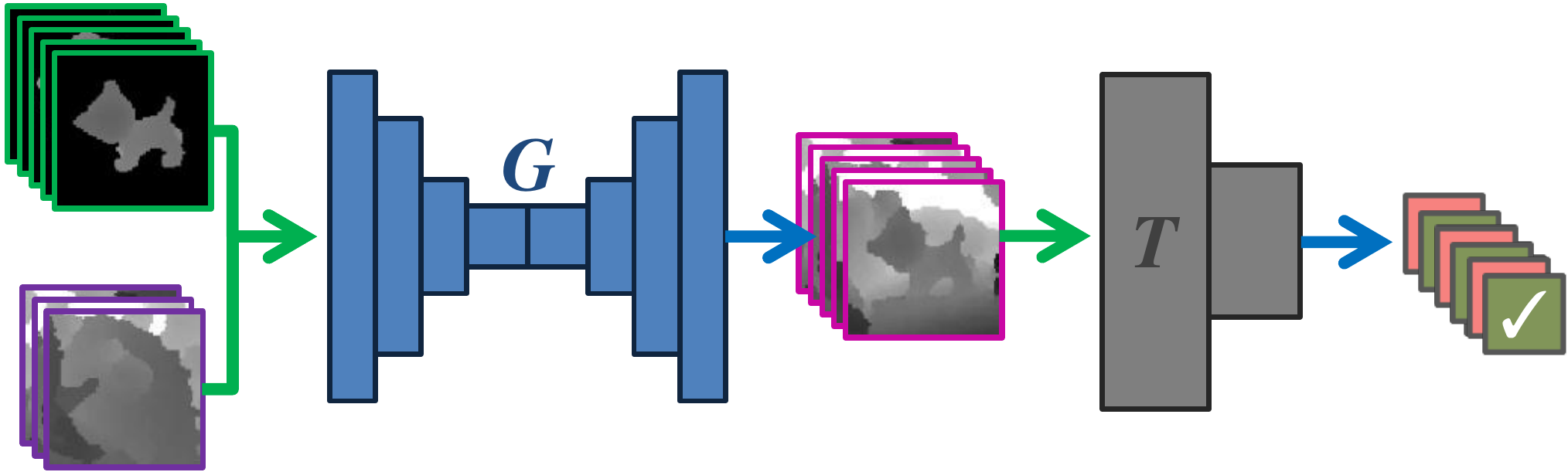}
      }
      &
      \parbox[c]{14em}{
      \includegraphics[width=\linewidth]{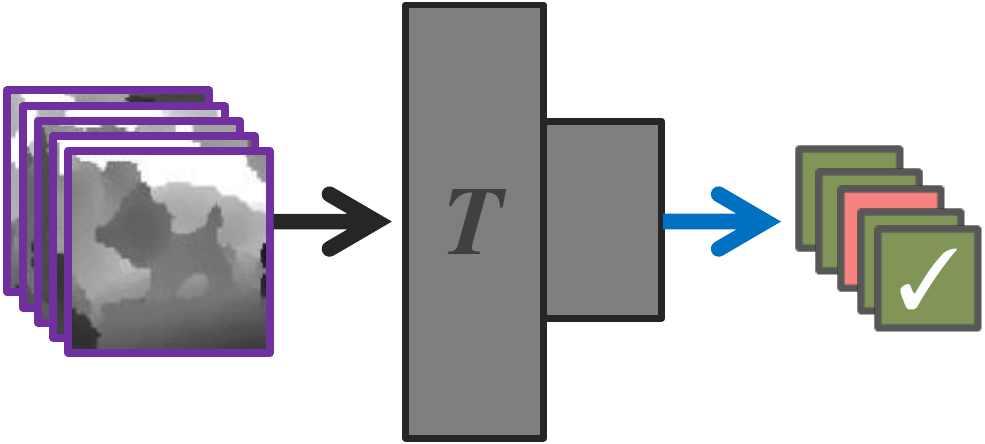}
      }
      \\[4.5em]
      
      & \parbox[c]{6mm}{\rotatebox[origin=c]{90}{\color{blueviolet}{\shortstack{Proposed \\ Approach}}}}  & 
      \parbox[c]{20.60em}{
      \includegraphics[width=\linewidth]{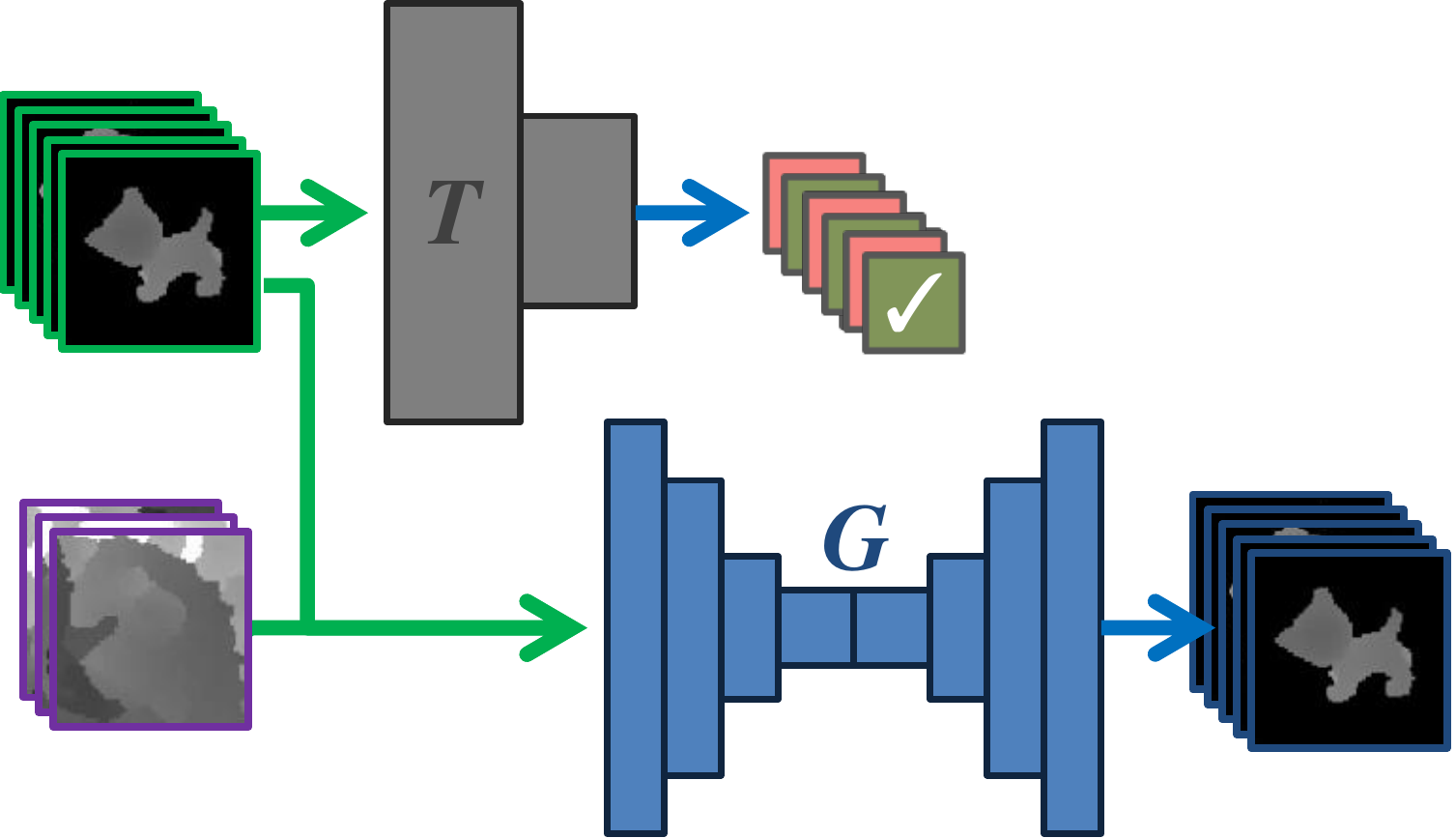}
      }
      &
      \parbox[c]{14em}{
      \includegraphics[width=\linewidth]{method_comp-real-test-da}
      }
      
      \\[4.5em]
      \midrule
   
  \parbox[c]{6mm}{\multirow{17}{*}{\rotatebox[origin=c]{90}{\textbf{\shortstack{\color{darkergreen}{Real Data Unavailable for Training}}}}}}
  
      & \parbox[c]{6mm}{\rotatebox[origin=c]{90}{\shortstack{Naive \\ Approach}}} & 
      \parbox[c]{14em}{
      \includegraphics[width=\linewidth]{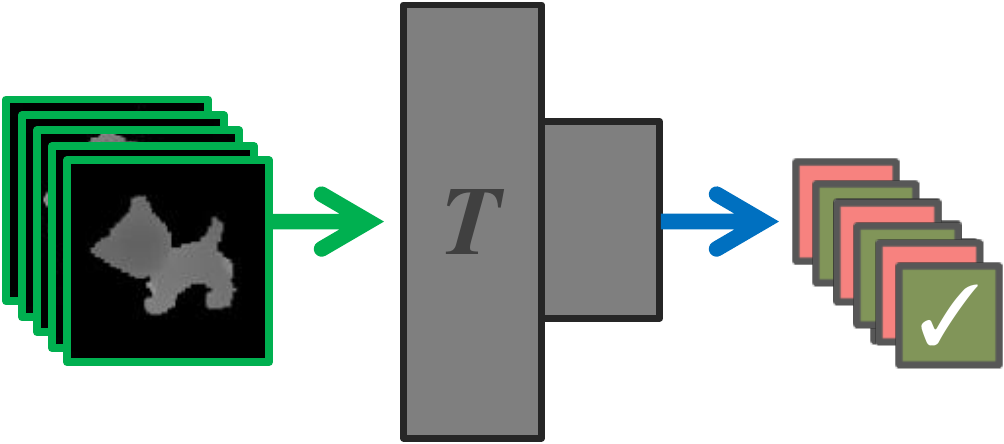}
      }
      &
      \parbox[c]{14em}{
      \includegraphics[width=\linewidth]{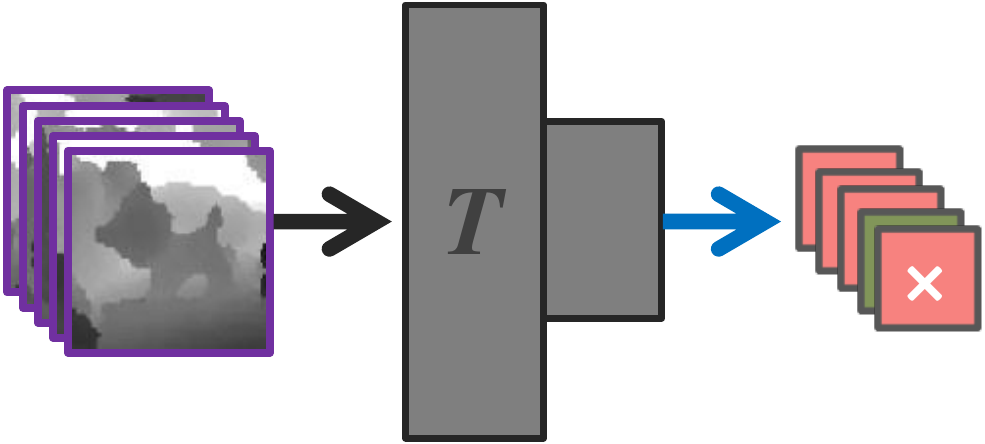}
      }
      
      \\[3.5em]
      
      & \parbox[c]{6mm}{\rotatebox[origin=c]{90}{\shortstack{Augmentation \\ Approaches  \cite{sadeghi2016cad,tobin2017domain}}}} & 
      \parbox[c]{24.23em}{
      \includegraphics[width=\linewidth]{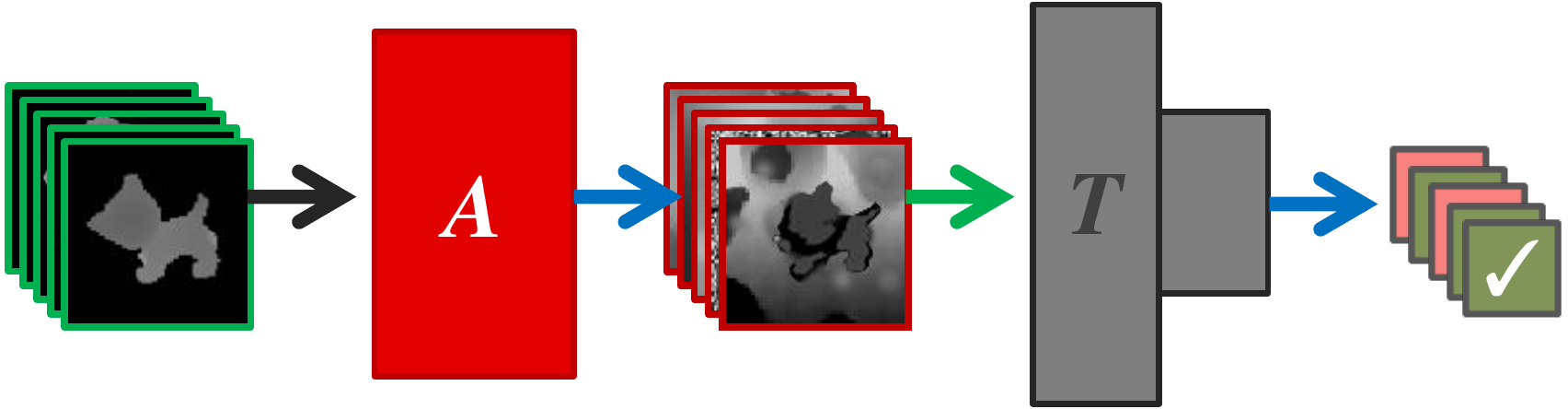}
      }
      &
      \parbox[c]{14em}{
      \includegraphics[width=\linewidth]{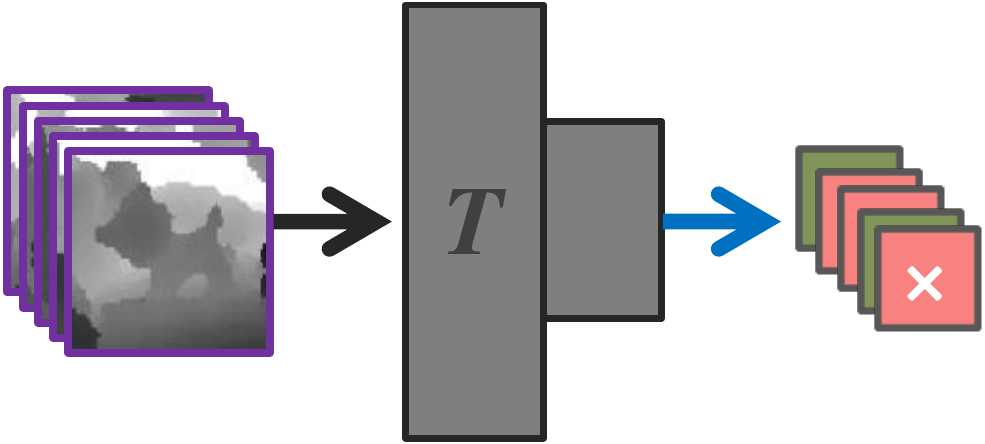}
      }
      
      \\[3.5em]
      
      & \parbox[c]{6mm}{\rotatebox[origin=c]{90}{\color{blue}{\shortstack{Proposed \\ Approach}}}} & 
      \parbox[c]{28em}{
      \includegraphics[width=\linewidth]{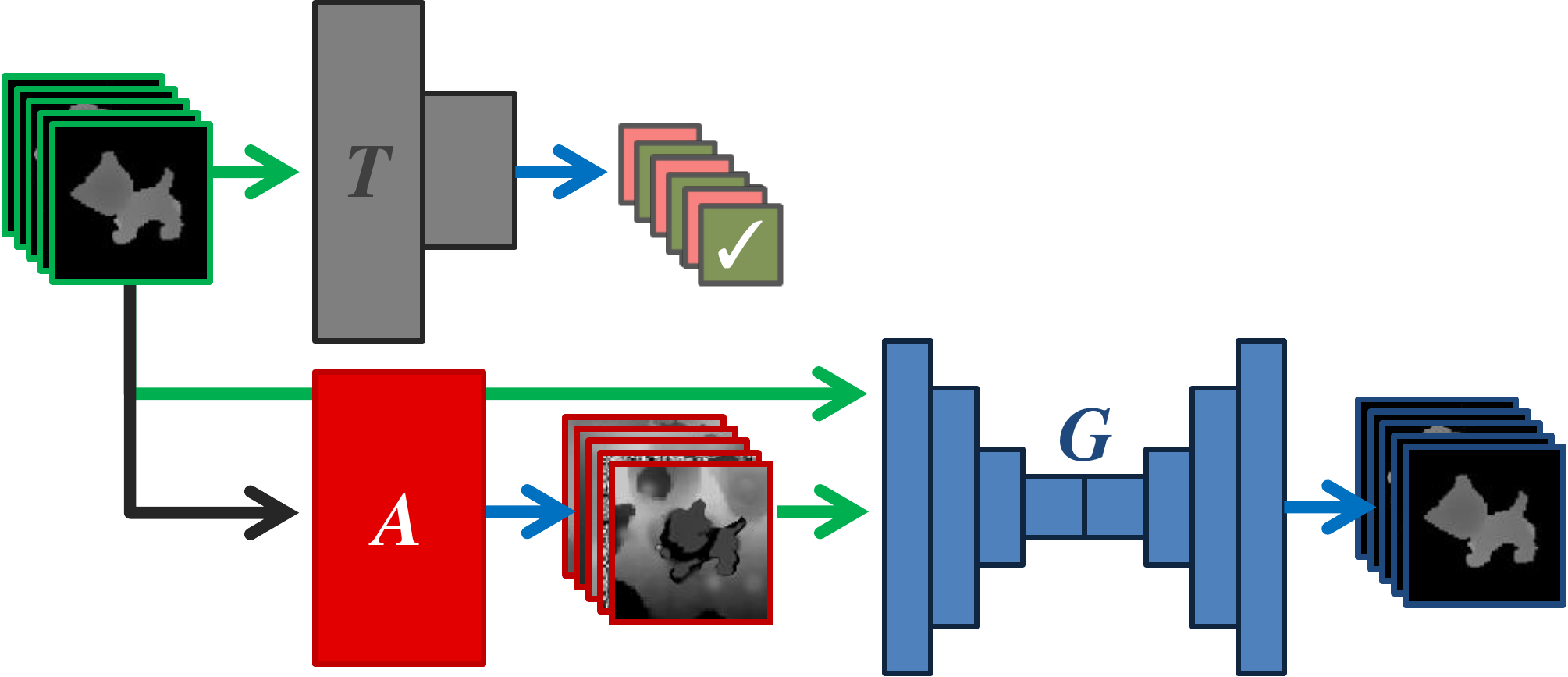}
      }
      &
      \parbox[c]{14em}{
      \includegraphics[width=\linewidth]{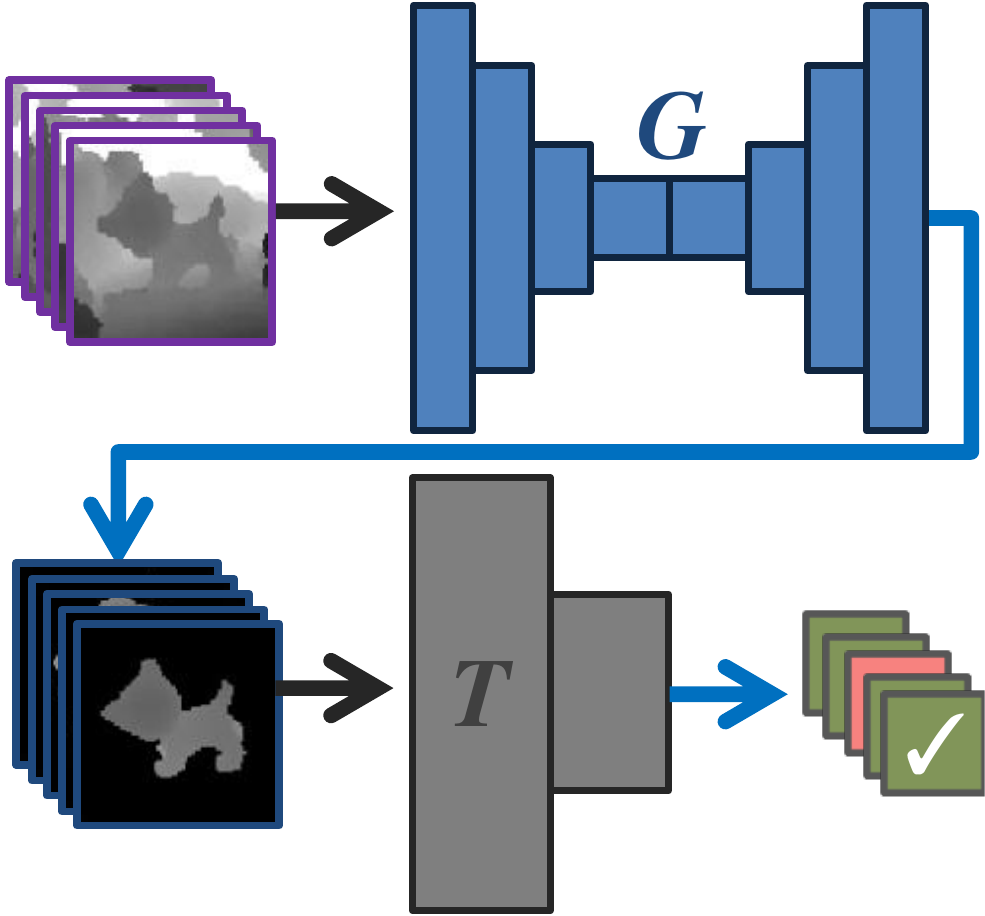}
      }
      
      \\[4.5em]
      \bottomrule
      \\[-1em]
      & & \multicolumn{2}{c}{
      \parbox[c]{43.4em}{
      \includegraphics[width=\linewidth]{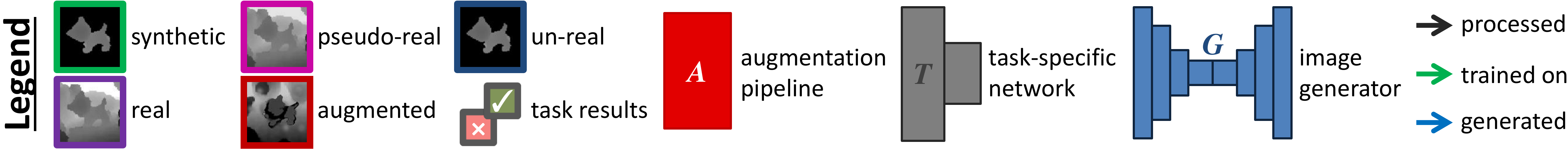}
      }
      }

  \end{tabular}
\end{table*}

% !TeX spellcheck = en_US

\begin{figure*}
  \centering
  \includegraphics[width=1\linewidth]{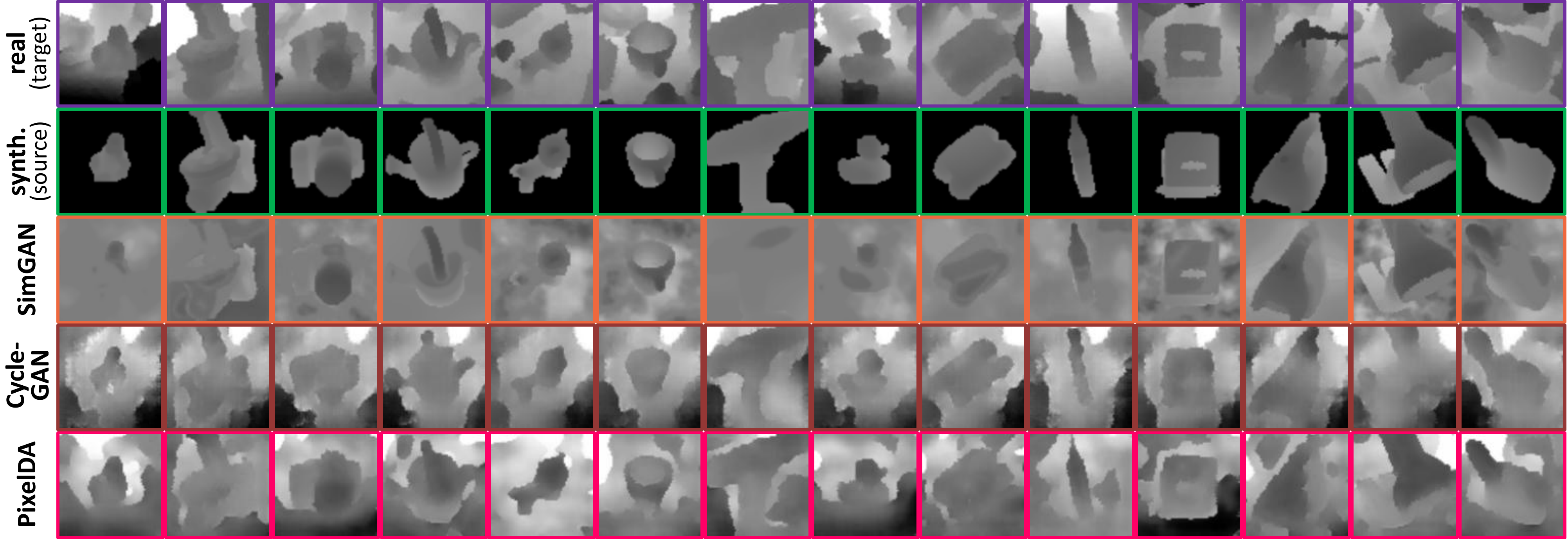}
  \caption{\textbf{Qualitative results of opposite domain adaptation GANs~\cite{shrivastava2016learning,zhu2017unpaired,bousmalis2016domain} on LineMOD~\cite{hinterstoisser2012model}}. First row contains indicative real images from the target domain; second row contains the synthetic depth images provided as sources; followed by the corresponding GANs outputs below.
  }
  \label{fig:gans_results}
\end{figure*}

\section{Additional Experiments and Comparisons}

\subsection{Qualitative Comparison of the Opposite Domain-Adaptation GANs}

To support the quantitative results and observations of the comparison between \textit{UnrealDA} and some state-of-the-art GAN-based domain-adaptation methods (in Section 4.3 of the paper), we provide a qualitative juxtaposition of results from SimGAN~\cite{shrivastava2016learning}, CycleGAN~\cite{zhu2017unpaired}, and PixelDA~\cite{bousmalis2016domain} for LineMOD~\cite{hinterstoisser2012model} in Figure~\ref{fig:gans_results}.

As discussed in the paper, we can note that despite filling the background with random noise to help the method, SimGAN's refiner fails to compensate for the missing information. 
CycleGAN does succeed at generating clutter, but at the expense of the foreground, which often ends up too distorted for recognition.
Thanks to its foreground loss and integrated classifier, PixelDA fares the best, though one can still notice artifacts and a slight lack of variability in the backgrounds.

\subsection{Scalability on BigBIRD Dataset}

\begin{figure*}
  \centering
  \includegraphics[width=1\linewidth]{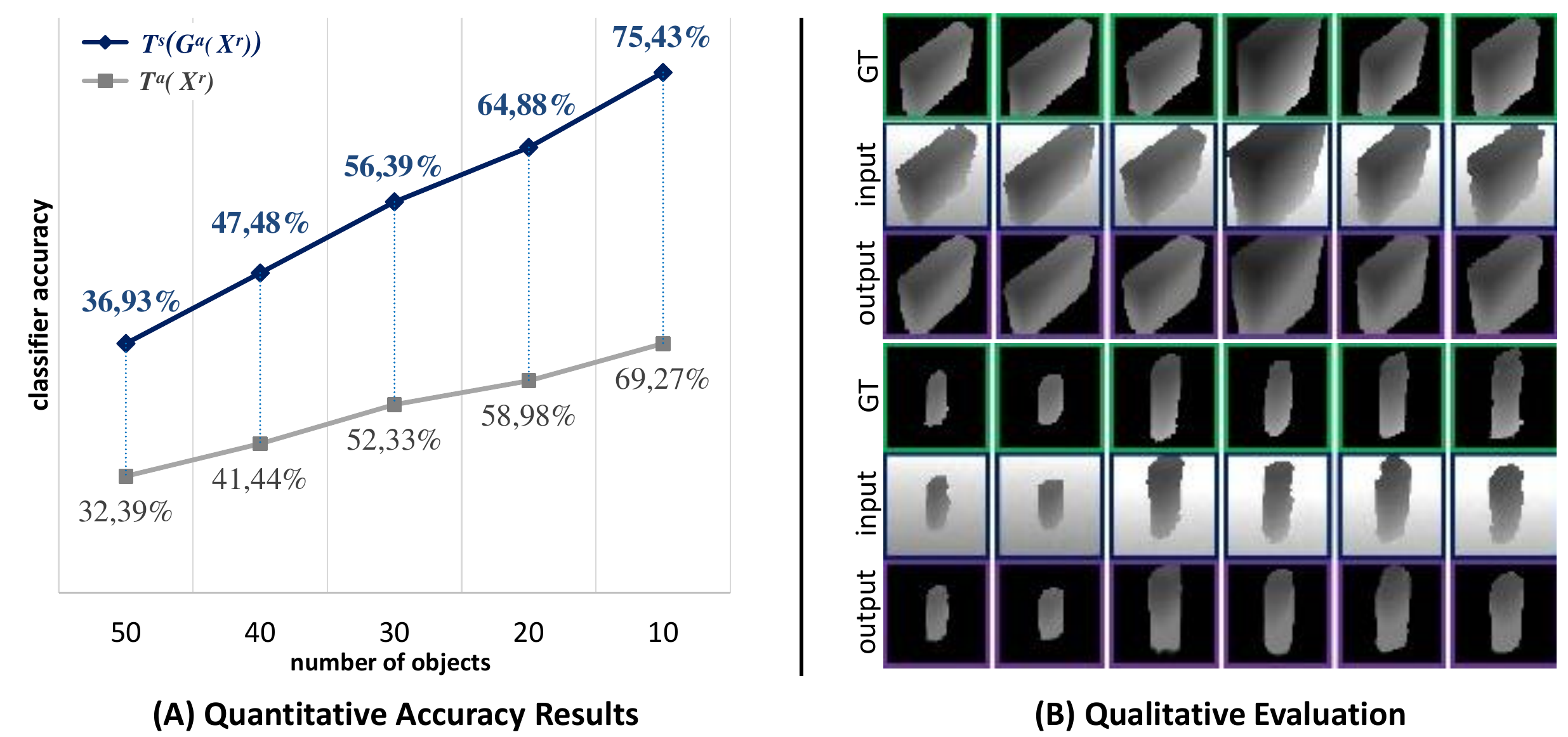}
  \caption{\textbf{(A) Quantitative results for a classification task on BigBIRD~\cite{singh2014bigbird}.} Comparing the performance of $T^a_{tri}$ on the real scans $X^r$, and $T^s_{tri}$ on $G^a(X^r)$ for different number of objects to classify (with $G^a$ constant, trained once for the 50 objects); 
  \textbf{(B) Qualitative results of $\pmb{G^a}$ pre-processing on the test set.} Note that every input image comes from a different class, highlighting the high inter-class similarities in this dataset. More images can be found in Figures~\ref{fig:sup_bigbird1} \&~\ref{fig:sup_bigbird2}  (\textit{GT} $=$ ground-truth).
 }
  \label{fig:bigbird_plot}
\end{figure*}

To demonstrate that our \textit{UnrealDA} method can pre-process images from a large number of classes, we proceed with an experiment on BigBIRD~\cite{singh2014bigbird}. 
BigBIRD is a dataset of RGB-D sequences and reconstructed 3D models of more than 100 objects. It is however extremely challenging, especially when only considering the depth modality for recognition (as pointed out in other works \eg~\cite{georgakis2017synthesizing,carlucci2018text}). The variability in terms of geometry is quite small (\eg boxes and bottles which cannot be distinguished among themselves without the color texture, \cf Figure~\ref{fig:bigbird_plot}-B), and more than twenty 3D models are corrupted (misconstructed because of the reflectivity of some materials \eg for glass and plastic bottles). The reflectivity of these objects and the turn-table itself (used to capture the dataset) also heavily impaired the quality of the test set, with vast portions of missing data in the scans.

For those reasons (and similarly to~\cite{georgakis2017synthesizing,carlucci2018text}), we selected a sub-set of 50 objects for 1) their clean 3D models (to generate our training data); 2) their relative geometrical variability (to make the pre-processing task more challenging for our \textit{UnrealDA} / to have a dataset more adequate to a depth-only experiment). Considering a single-view classification task on these scans using the triplet method $T_{tri}$ (\cf experiments in paper), we applied the following experimental protocol to isolate the contributions of \textit{UnrealDA} and demonstrate that even for a larger number of classes, our pre-processing consistently improves recognition.

Assuming no real data available, we train a single instance $G^a$ of our \textit{UnrealDA} pipeline on synthetic augmented images of the 50 objects. We then train several instances of $T_{tri}$ to classify among an increasingly larger number of objects (\ie one $T_{tri}$ trained to classify a subset of 10 objects, one for a subset of 20, \etc up to 50). For each sub-task, we compare the performance of $T^a_{tri}$ on the real scans $X^r$, and $T^s_{tri}$ on $G^a(X^r)$. Results are presented in Figure~\ref{fig:bigbird_plot}.

As expected, we can observe that the accuracy of $T_{tri}$ decreases when the number of (rather ambiguous) classes increases. However, the performance boost provided by our $UnrealDA$ trained on all 50 objects stays almost constant, with an increase of {\raise.17ex\hbox{$\scriptstyle\sim$}}$5\%$ in classification accuracy compared to the $T^a_{tri}$ modality.

% !TeX spellcheck = en_US

	\section{Implementation Details}
	\subsection{GAN Architecture and Parameters} 
\begin{figure*}
  \centering
  \includegraphics[width=1\linewidth]{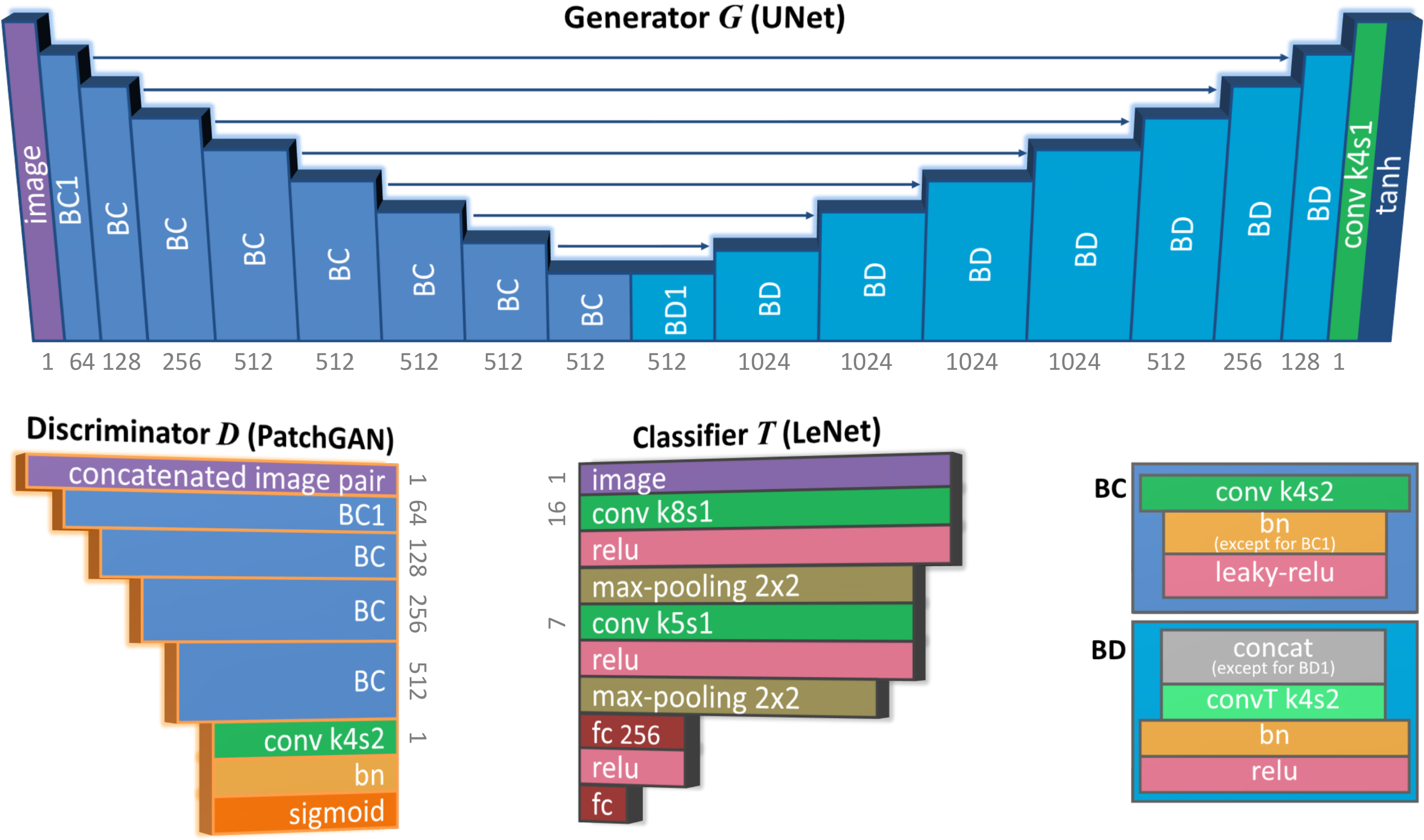}
  \caption{\textbf{Overview of the networks architectures}. \textit{conv k4s2} stands for a convolutional layer with $4 \times 4$ filters and a stride of 2, \textit{convT} stands for a transposed convolution, \textit{bn} for a batch-normalization layer, \textit{fc} for a fully-connected one. Arrows represent the skip connections between blocks.
  }
  \label{fig:net-architecture}
\end{figure*}

	As mentioned in Subsection~\refwithdefault{sec:archi}{3.4} of the paper: 
	though our solution is not bound to particular network architectures, we opted for the work of Isola~\etal~\cite{isola2016image}, given its efficiency and popularity at the time of this paper.
	As shown in Figure~\ref{fig:net-architecture}, the generator $G$ is an adaptation of the U-Net architecture~\cite{ronneberger2015u}, and the discriminator of the $70 \times 70$ \textit{PatchGAN} architecture~\cite{isola2016image}. 
	
	Let $B^C_k$ be an encoding block made of the following layers: 2-factor downsampling Convolution with $k$ filters, BatchNorm, then LeakyReLU. 
	Let $B^D_k$ be a decoding block made of: 2-factor upsampling Transposed Convolution with $k$ filters, BatchNorm, LeakyReLU, then Dropout layers.
	
	\par\noindent
	\textbf{Generator $\mathbf{G}$:} The encoding part of the generator is made of $B^C_{64}-B^C_{128}-B^C_{256}-B^C_{512}-B^C_{512}-B^C_{512}-B^C_{512}-B^C_{512}$, and its decoding part of $B^D_{512}-B^D_{1024}-B^D_{1024}-B^D_{1024}-B^D_{1024}-B^D_{512}-B^D_{256}-B^D_{128}$. Each encoding block has a skip connection toward the opposite decoding block to concatenate the channels.
	
	\par\noindent
	\textbf{Discriminator $\mathbf{D}$:} It is a simple CNN made of $B^C_{64}-B^C_{128}-B^C_{256}-B^C_{512}$, followed by a last convolution layer mapping to an one-dimensional output, and by a sigmoid function layer~\cite{isola2016image}.
	
	All networks are implemented using the TensorFlow framework~\cite{abadi2016tensorflow} in Python.
	
	\par\noindent
	\textbf{Hyper-parameters:} The network architectures are further defined by the following parameters:
	\begin{itemize}
		\setlength\itemsep{0em}
		\item All Convolution layers have $4 \times 4$ filter kernels;
		\item All Dropout layers have a dropout rate of $50\%$;
		\item All LeakyReLU layers have a leakiness of $0.2$;
		\item All depth images (input and output) are single-layer $64 \times 64$px patches, normalized between 0 and 1;
		\item The batch size is set to $1$.\footnote{Batch normalization with a batch size of 1 is known as ``instance normalization" and seems to be effective for image generation~\cite{isola2016image,ulyanov2016instance}}
	\end{itemize}
	
	Training parameters are:
	\begin{itemize}
		\setlength\itemsep{0em}
		\item Weights are initialized from a zero-centered Gaussian with a standard deviation of $0.02$;
		\item The Adam optimizer~\cite{kingma2014adam} is used, with $\beta_1 = 0.5$;
		\item The base learning rate is set at $0.0002$.
	\end{itemize}
	
	Finally, the different loss components are weighted as follows in our experiments:
	\begin{itemize}
		\setlength\itemsep{0em}
		\item The discriminative loss $\mathcal{L}_{d}$ is weighted by $\alpha = 1$;
		\item The L1 similarity loss $\mathcal{L}_{g}$ is weighted by $\beta = 100$;
		\item The foreground loss $\mathcal{L}_{f}$ is weighted by $\delta = 200$;
		\item The task-specific loss $\mathcal{L}_{t}$ is weighted by $\gamma = 10$ when used.
	\end{itemize}

	\subsection{Augmentation Pipeline Details}
	
	\par\noindent
	\textbf{Background noise:}
	All the noise types used for our augmentation pipeline are generated using the open-source FastNoise library~\cite{fastnoise}. In particular, three noise modalities provided by the framework are used: Perlin noise~\cite{perlin2002improving}, cellular noise~\cite{worley1996cellular}, and white noise. Noise frequencies for all modalities are sampled from the uniform distribution $\mathcal{U}(0.0001,0.1)$.

	\begin{algorithm}
		\KwIn{ 
			$x \in \mathbb{R}^{l_x \times l_y}$ depth image,
			$z \in \mathbb{Z}^k$ noise vector
		}
		\KwOut{ 
			$x_a \in \mathbb{R}^{l_x \times l_y}$ augmented depth image	
		}
		
		\tcc{sampling distortion parameters from the noise vector:}
		$f_{X}, f_{Y}, f_{Z}, w_{XY}, w_{Z}  \leftarrow \text{sampleFromVector}(z)$\;
		\tcc{generating 3D vector field:}
		$v_d[0] \leftarrow \text{fast2DNoise}(f_{X})$\;
		$v_d[1] \leftarrow \text{fast2DNoise}(f_{Y})$\;
		$v_d[2] \leftarrow \text{fast2DNoise}(f_{Z})$\;
		\tcc{applying 3D distortion:}
		\For{$ i \in \{1, \ldots, l_x\} $}{
			\For{$ j \in \{1, \ldots, l_y\} $}{
				$\begin{aligned}
				x_a(i,j) \leftarrow  x(&i + w_{XY} * v_d[0][i,j], \\
				&j + w_{XY} * v_d[1](i,j) ) \\
				& + w_{Z} * v_d[2](i,j)
				\end{aligned}$
				
			}
		}
		\KwRet{$x_a$}
		
		\caption{Foreground distortion}
		\label{alg:foreground}
	\end{algorithm}
	\par\noindent
	\textbf{Foreground distortion:} 
	The foreground distortion component warps the input image using a random vector field. As a first step, three Perlin noise images are generated using the FastNoise library~\cite{fastnoise} to form a three-dimensional offset vector field $v_d$. The first two dimensions are used for depth value distortions in X and Y axes of the image space, whereas the third dimension is applied to the Z image depth values directly. Since values of $v_d$ vary between $-1$ and $1$, warping factors $w \in \mathbb{R}^3$ that control the distortion effect are introduced. Once the 3D offset vector generated, it is applied to the input image $x$ to generate the output image $x_a$. The pseudocode is presented in Algorithm \ref{alg:foreground}. 
	
	In our pipeline, the noise frequencies as well as the warping factors are defined by the randomly sampled vector $z$. More precisely, for the presented experiments, $f_{X}$ and $f_{Y}$ are sampled at each iteration from the uniform distribution $\mathcal{U}(0.0001,0.1)$,  $f_{Z}$ from $\mathcal{U}(0.01,0.1)$ (higher frequencies), $w_{XY}$ from $\mathcal{U}(0, 10)$, and $w_{Z}$ from $\mathcal{U}(0, 0.005)$.

	\par\noindent
	\textbf{Random occlusions:}
	Our occlusion generation function is based on the algorithm used in \cite{ounsworth2015anticipatory}, where they generate 2D obstacles of various complexity levels applied to a drone moving planning simulation. 
	As mentioned in Subsection~\refwithdefault{sec:augmentation}{3.3} of the paper, the main idea is to sample points by walking around the circle taking random angular steps and random radii at each step. When points $p$ are generated, they are used to form a polygon filled with depth values varying between 0 and the shortest distance to the object's surface. The procedure can be repeated several times depending on the desired occlusion level.

The complexity of each polygon is defined by parameters $\epsilon$ ("irregularity"), which sets an error to the default uniform angular distribution, and $\sigma$ ("spikeyness"), which controls how much point coordinates vary from the radius $r_{ave}$. The pseudocode is listed in Algorithm~\ref{alg:occlusion}. 

\begin{algorithm}[h]
	\KwIn{ 
			$z \in \mathbb{Z}^k$ noise vector
		}
		\KwOut{ 
			$p = \{p_i \in \mathbb{R}^2\}_{i=0}^{N_{vert}}$ polygon points	
		}
		
		\tcc{sampling occlusion parameters from the noise vector:}
		$c_{x}, c_{y}, r_{ave}, N_{vert}, \epsilon , \sigma  \leftarrow \text{sampleFromVector2}(z)$\;

		\tcc{generating angle steps:}
		%	$\theta \leftarrow \{\}$ \;
		$sum = 0$ \;
		\For{$ i \in \{1, \ldots, N_{vert}\} $}{
			$\delta\theta_{i} \leftarrow \mathcal{U}(\sfrac{2 \pi}{N_{vert}} - \epsilon, \sfrac{2 \pi}{N_{vert}} + \epsilon)$ \;
			%		$\delta\theta \leftarrow \delta\theta \bigcup \{\delta\theta_{tmp}\}$ \;
			$sum \leftarrow sum + step$ \;
		}
		\tcc{normalizing the steps:}
		$k \leftarrow \sfrac{sum}{(2 \pi)}$ \;
		\For{$ i \in \{1, \ldots, N_{vert}\}$}{
			$\delta\theta_i \leftarrow \sfrac{\delta\theta_i}{k}$ \;
		}
		\tcc{generating polygon points:}
		%	$p \leftarrow \{\}$ \;
		$\theta_1 \leftarrow \mathcal{U}(0, 2 \pi)$ \;
		\For{$ i \in \{1, \ldots, N_{vert}\} $}{
			$r \leftarrow \mathcal{N}(r_{ave}, \sigma)$ \;
			$p_i \leftarrow (c_X + r \cos(\theta_i), c_Y + r \sin(\theta_i))$ \;
			$\theta_i \leftarrow \theta_i + \delta\theta_i$
		}
		\KwRet{p}
		
		\caption{Random polygon generation \cite{ounsworth2015anticipatory}}
		\label{alg:occlusion}
	\end{algorithm}

	Variables $c_X$ and $c_Y$ define the polygon center; $r_{ave}$ its average radius; $\delta \theta$ and $\theta$ are a vector of angle steps, and a vector of angles respectively; and $l_x \times l_y$ are the image dimensions (equal to $64  \times 64$ px here).  
	
	As for every augmentation step, each of these occlusion parameters is defined by the noise vector $z$. For our experiments, we chose the following sampling distributions (with $\mathcal{B}$ -- Bernoulli, $\mathcal{U}$ -- Uniform, and $\mathcal{N}$ -- Gaussian): 
\begin{itemize}
\setlength\itemsep{0em}
\item $\mathcal{B}\big(\mathcal{U}(0, \sfrac{l_x}{4}),\,\mathcal{U}(\sfrac{l_x}{4}, l)\big)$ for $c_X$;
\item $\mathcal{B}\big(\mathcal{U}(0, \sfrac{l_y}{4}),\,\mathcal{U}(\sfrac{l_y}{4}, l)\big)$ for $c_Y$;
\item $\mathcal{U}(10, \sfrac{l}{4})$ for $r_{ave}$, with $l = \min(l_x, l_y)$;
\item $\mathcal{U}(3, 10)$ for $N_{vert}$;
\item $\mathcal{U}(0, 0.5)$ for $\sigma$.
\end{itemize}

	\subsection{Depth Image Rendering}
	
	As explained in Subsections~\refwithdefault{sec:data_generation}{3.2} and~\refwithdefault{sec:datasets}{4.1} of the paper, the generation of noiseless synthetic depth images from the 3D models was simply done using OpenGL~\cite{opengl}, printing the \textit{z-buffer} content into an image for every viewpoint. Viewpoints were sampled by selecting the vertices of an icosahedron centered on the target objects. Finer sampling can be achieved by subdividing each triangular face of the icosahedron into four smaller triangles; and in-plane rotations can be added by rotating the camera around the axis pointing to the object for each vertex.
	
\begin{figure}[!htbp]
		\centering
		
		\subfigure[Axis symmetric]
			{\includegraphics[width=.32\linewidth]{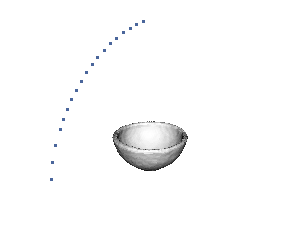}\label{fig:axis}}
		 \hfill
		\subfigure[Plane symmetric]
			{\includegraphics[width=.32\linewidth]{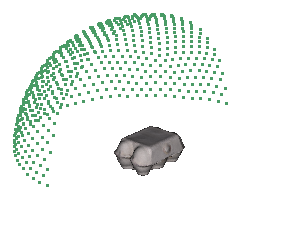}\label{fig:plans}}
		 \hfill
		\subfigure[Regular]
			{\includegraphics[width=.32\linewidth]{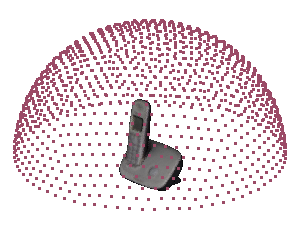}\label{fig:regular}}

		\caption{\textbf{Vertices sampling for different objects types of LineMOD objects} -- each vertex represents a camera position from which the object is rendered.}
		\label{fig:samplings}
	\end{figure}

	For the LineMOD~\cite{hinterstoisser2012model} synthetic dataset, we chose an icosahedron of radius 600mm with 3 consecutive subdivisions, and we considered only the vertices of its upper half (\ie no images shot from below the object). Seven in-plane rotations were added for each vertex, from -$45^\circ$ to $45^\circ$ with a stride of $15^\circ$. 
	Rotation invariance of four out of fifteen (\textit{cup}, \textit{bowl}, \textit{glue}, and \textit{eggbox}) LineMOD objects was also taken into account by limiting the amount of sampling points, such that each patch is unique. Figure~\ref{fig:samplings}  demonstrates the results of the output vertex sampling for different object types. We thus generated 2359 depth images for each of the 11 regular objects, 1239 for 3 plane symmetric objects (\textit{cup}, \textit{glue}, and \textit{eggbox}) and 119 for the axis symmetric \textit{bowl\textit{}}. 
%	
%	\begin{figure}[!htbp]
%		\centering
%		\begin{subfigure}[b]{.32\linewidth}
%			\centering
%			\includegraphics[width=\linewidth]{bowl}
%			\caption{Axis symmetric}
%			\label{fig:axis}
%		\end{subfigure} \hfill
%		\begin{subfigure}[b]{.32\linewidth}
%			\centering
%			\includegraphics[width=\linewidth]{eggbox}
%			\caption{Plane symmetric}
%			\label{fig:plans}
%		\end{subfigure} \hfill
%		\begin{subfigure}[b]{.32\linewidth}
%			\centering
%			\includegraphics[width=\linewidth]{phone}
%			\caption{Regular}
%			\label{fig:regular}
%		\end{subfigure}
%		\caption{\textbf{Vertices sampling for different objects types of LineMOD objects} -- each vertex represents a camera position from which the object is rendered.}
%		\label{fig:samplings}
%	\end{figure}
	
	Similarly for each of the 50 objects selected from BigBIRD~\cite{singh2014bigbird}, we considered the upper half of an icosahedron subdivided 3 times but with no in-plane rotations (as the test set contains none).
	
	For T-LESS~\cite{hodan2017t} synthetic data, the whole icosahedron, with a radius of 600mm and 2 subdivisions, was considered (since objects in this dataset can lie on different sides from one scene to another). Since the test dataset contains no in-plane rotations, none were added to the training data neither.
	This led to the generation of 162 depth images per object (for 5 objects --- numbers 2, 6, 7, 25, and 29).

	\section{Additional Qualitative Results}
	
\begin{figure*}[h!]
		\centering
		\includegraphics[width=.9397\linewidth]{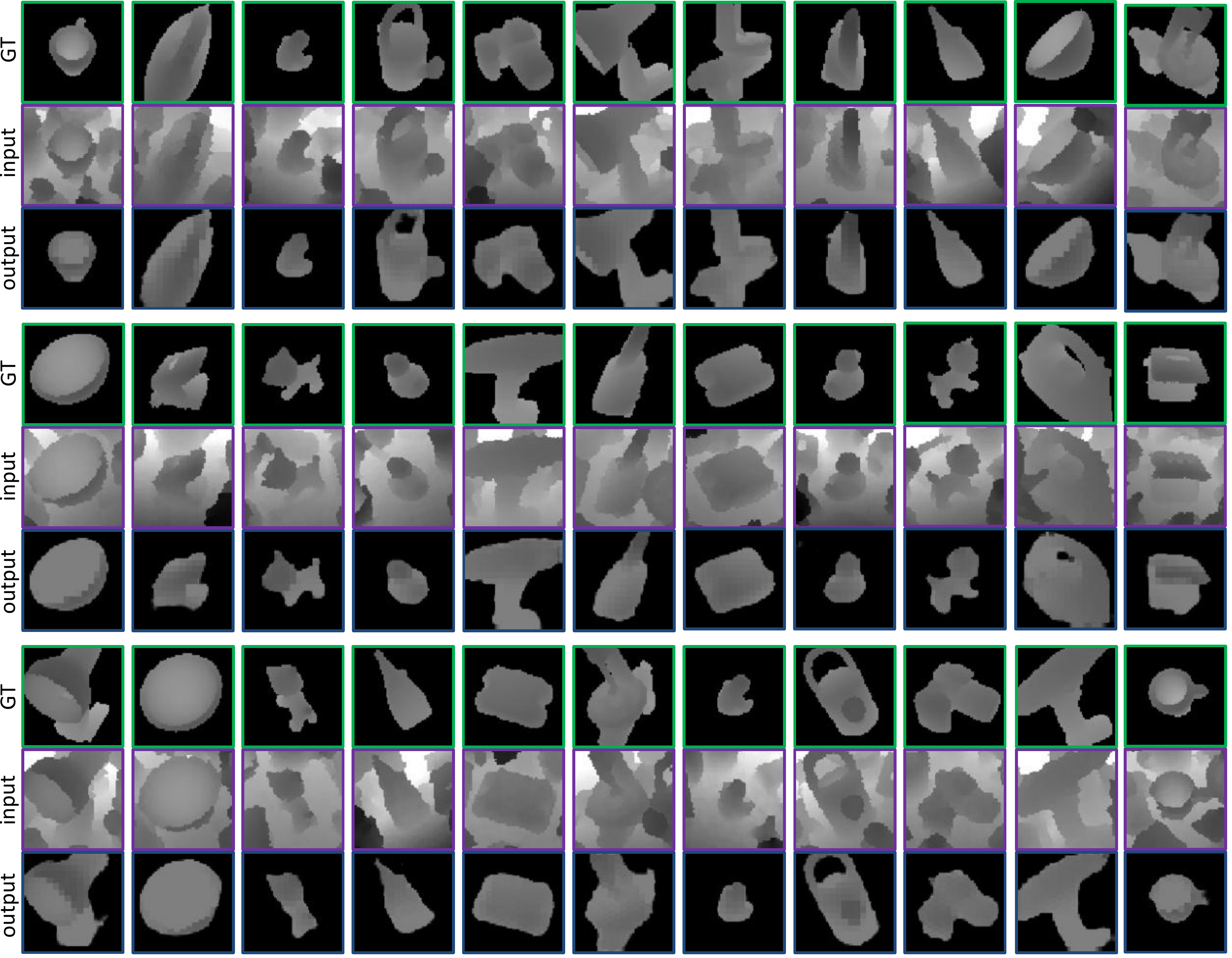}
		\caption{\textbf{Results on LineMOD}~\cite{hinterstoisser2012model} -- real scans (\textit{input}) processed by our method, compared to the ground-truth (\textit{GT}).}
		\label{fig:sup_linemod}  
	\end{figure*}

\begin{figure*}[h!]
		\centering
		\includegraphics[width=.9397\linewidth]{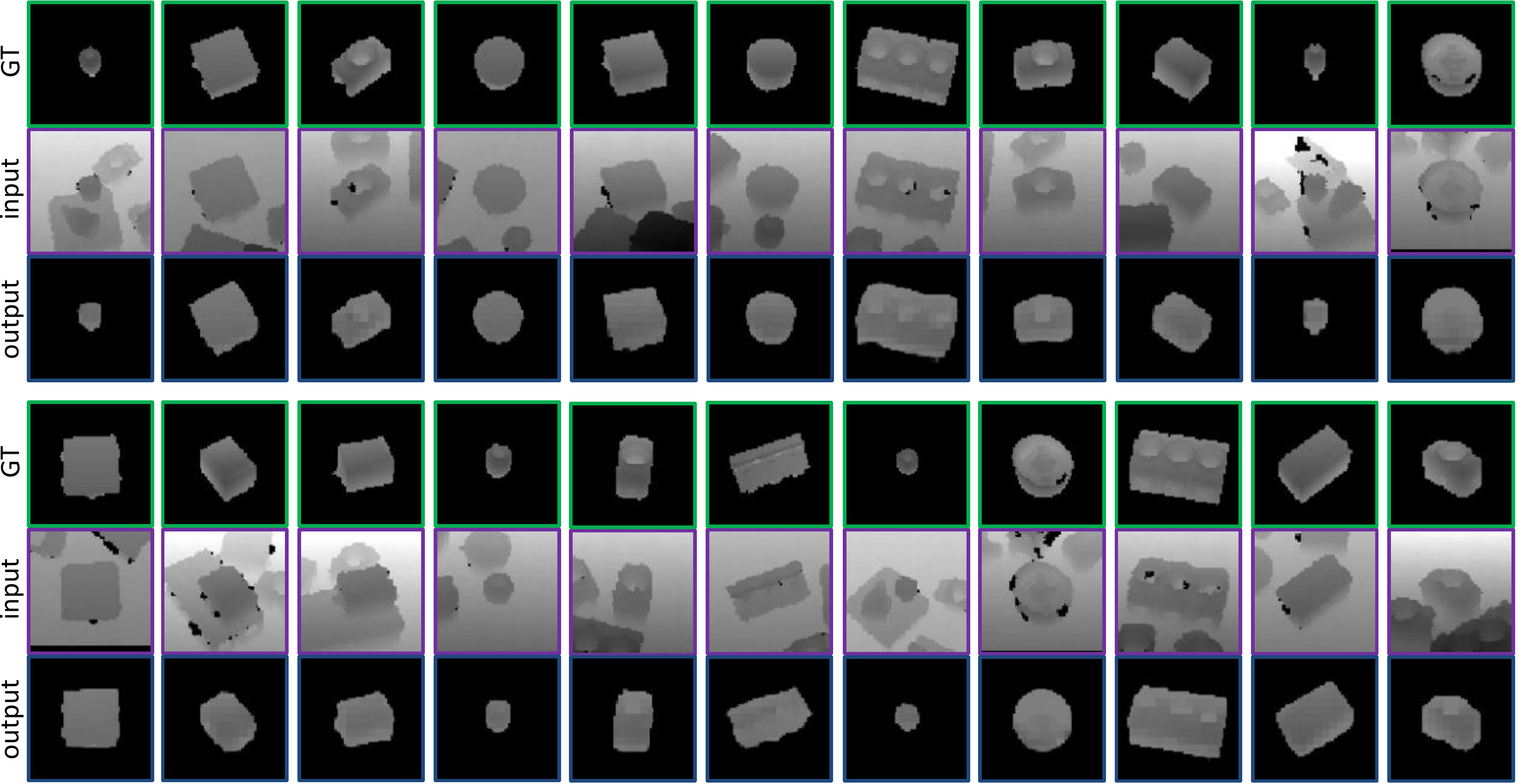}
		\caption{\textbf{Results on T-LESS}~\cite{hodan2017t} -- real scans (\textit{input}) processed by our method, compared to the ground-truth (\textit{GT}).}
		\label{fig:sup_tless}  
	\end{figure*}

\begin{figure*}[h!]
		\centering
		\includegraphics[width=1\linewidth]{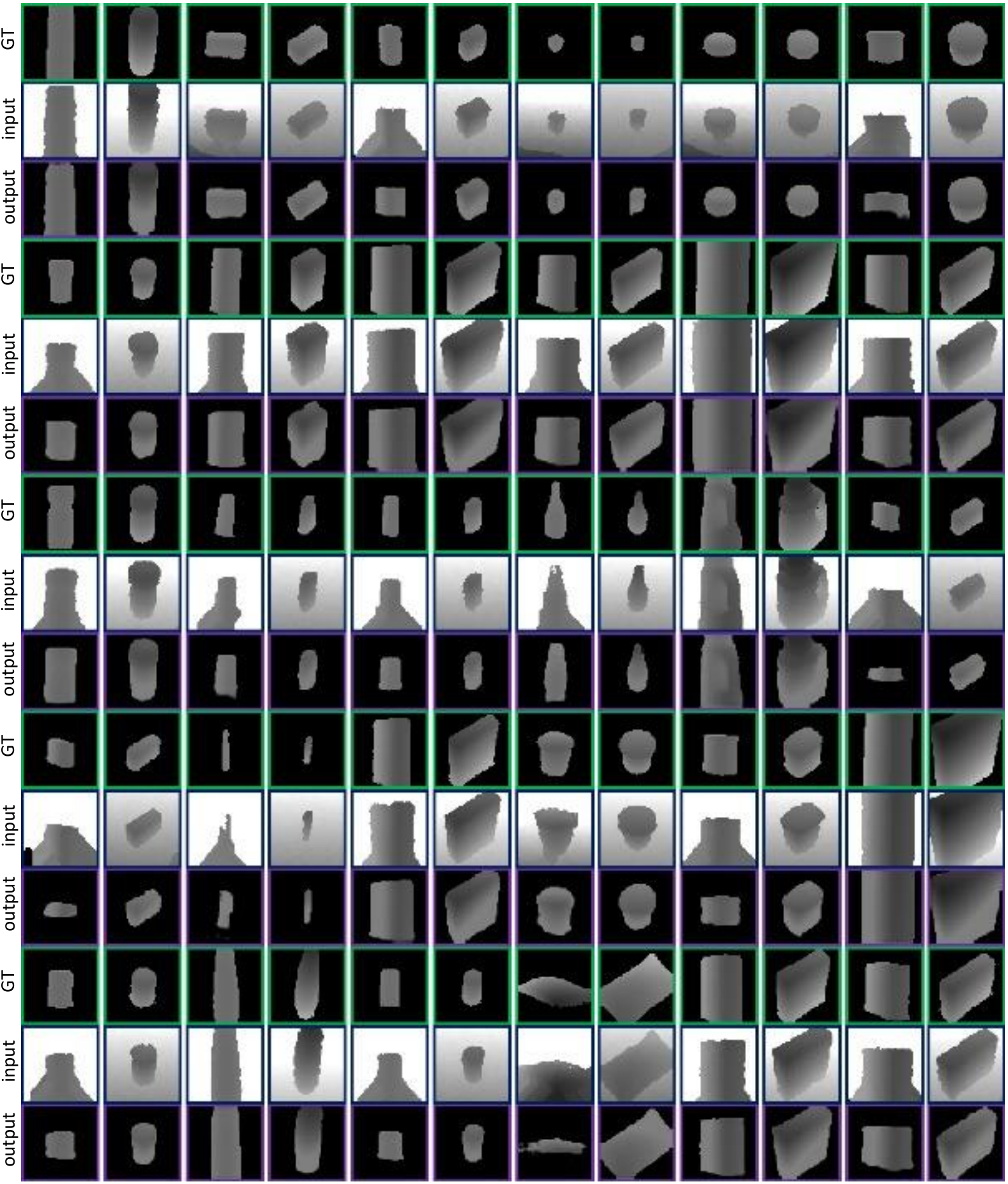}
		\caption{\textbf{Results on BigBIRD~\cite{singh2014bigbird} (1/2)} -- real scans (\textit{input}) processed by our method, compared to the ground-truth (\textit{GT}). Each of the 50 selected objects is displayed twice with different poses.}
		\label{fig:sup_bigbird1}  
	\end{figure*}

\begin{figure*}[t]
		\centering
		\includegraphics[width=1\linewidth]{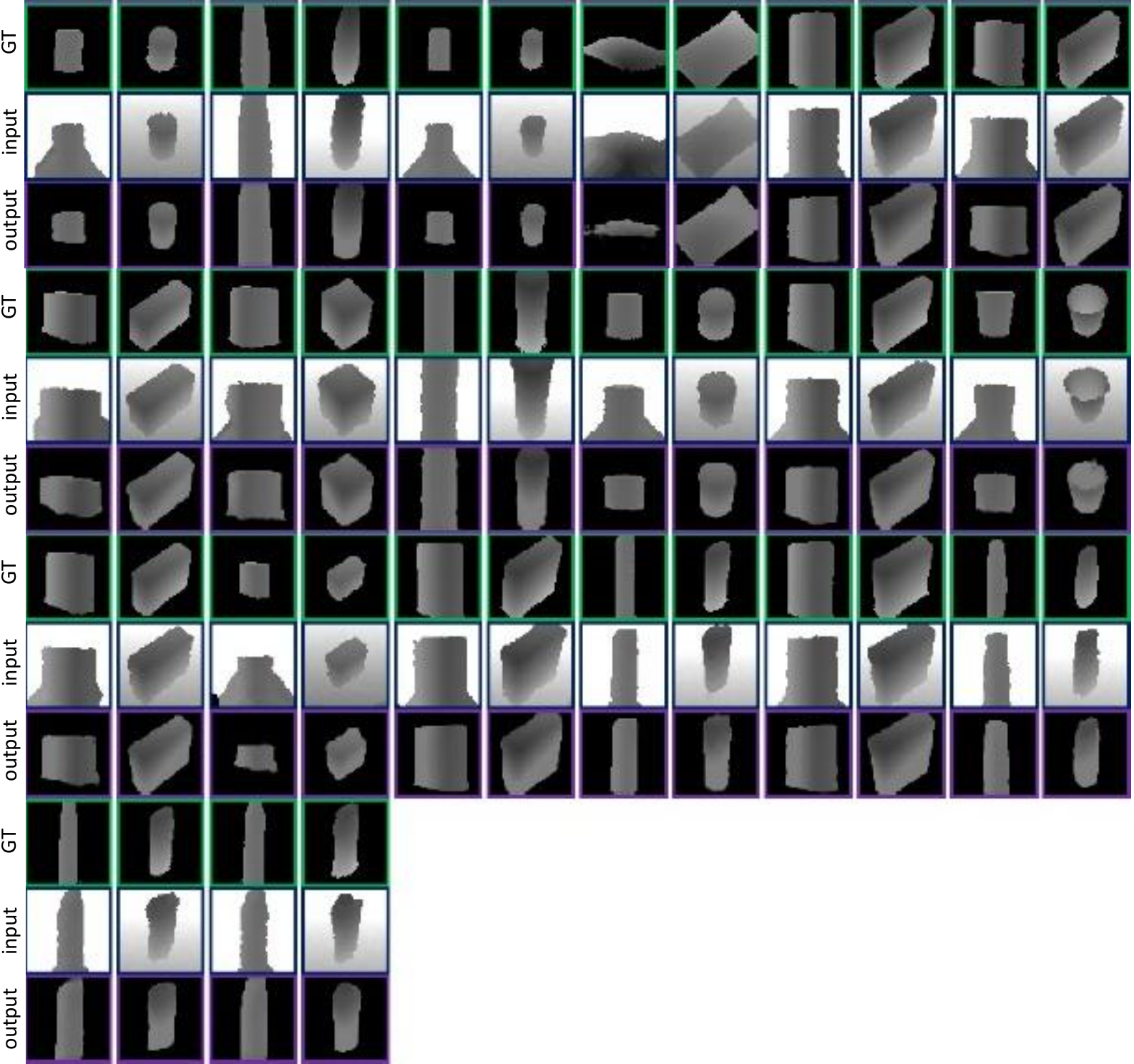}
		\caption{\textbf{Results on BigBIRD~\cite{singh2014bigbird} (2/2)} -- real scans (\textit{input}) processed by our method, compared to the ground-truth (\textit{GT}). Each of the 50 selected objects is displayed twice with different poses.}
		\label{fig:sup_bigbird2}  
	\end{figure*}

	Figures~\ref{fig:sup_linemod},~\ref{fig:sup_tless}, ~\ref{fig:sup_bigbird1}, and~\ref{fig:sup_bigbird2} contain further visual results, for the processing by our pipeline of LineMOD~\cite{hinterstoisser2012model}, T-LESS~\cite{hodan2017t}, and BigBIRD~\cite{singh2014bigbird} real depth images.
	
\clearpage
\clearpage

{\small
\bibliographystyle{ieee}
\bibliography{../references}
}

\end{document}